\documentclass[journal]{IEEEtai}  
\usepackage[colorlinks,urlcolor=blue,linkcolor=blue,citecolor=blue]{hyperref}

\usepackage{adjustbox}
\usepackage{colortbl}

\usepackage[dvipsnames]{xcolor}
\definecolor{lightgreen}{RGB}{173,213,176}
\definecolor{lightyellow}{RGB}{255,255,204}
\definecolor{lightblue}{RGB}{204,255,255}
\definecolor{lightred}{RGB}{255,204,204}

\usepackage{pifont}
\newcommand{\cmark}{\ding{51}}%
\newcommand{\xmark}{\ding{55}}%
\usepackage{svg}
\usepackage[dvipsnames]{xcolor}
\usepackage{graphicx} 
\usepackage{tabularx}

\graphicspath{{figures-final/}} 
\DeclareGraphicsExtensions{.pdf,.jpg,.png} 
\usepackage{amsmath,bm} 
\usepackage{amsfonts} 
\usepackage{mathtools} 
\usepackage{color} 
\usepackage{multirow} 
\usepackage{hhline} 
\usepackage{enumerate} 
\usepackage{epstopdf} 
\usepackage{dsfont}
\usepackage{siunitx} 
\usepackage{cite}   
\usepackage{diagbox}    
\usepackage{pdfpages}   
\usepackage{breqn}
\usepackage[caption=false]{subfig}
\usepackage{comment}
\colorlet{Green1}{green!90!}
\colorlet{Green2}{green!60!}
\colorlet{Green3}{green!40!}
\colorlet{Green4}{green!20!}
\colorlet{Green5}{green!10!}

\definecolor{Bookcolor}{HTML}{00F9DE}
\definecolor{darkgreen}{rgb}{0.0, 0.5, 0.0}

\newcommand{\argmin}[1]{\underset{#1}{\operatorname{arg}\,\operatorname{min}}\;}

\usepackage{scalerel} 
\usepackage{tikz} 
\usetikzlibrary{svg.path} 
\definecolor{orcidlogocol}{HTML}{A6CE39}
\tikzset{
  orcidlogo/.pic={
    \fill[orcidlogocol] svg{M256,128c0,70.7-57.3,128-128,128C57.3,256,0,198.7,0,128C0,57.3,57.3,0,128,0C198.7,0,256,57.3,256,128z};
    \fill[white] svg{M86.3,186.2H70.9V79.1h15.4v48.4V186.2z}
                 svg{M108.9,79.1h41.6c39.6,0,57,28.3,57,53.6c0,27.5-21.5,53.6-56.8,53.6h-41.8V79.1z M124.3,172.4h24.5c34.9,0,42.9-26.5,42.9-39.7c0-21.5-13.7-39.7-43.7-39.7h-23.7V172.4z}
                 svg{M88.7,56.8c0,5.5-4.5,10.1-10.1,10.1c-5.6,0-10.1-4.6-10.1-10.1c0-5.6,4.5-10.1,10.1-10.1C84.2,46.7,88.7,51.3,88.7,56.8z};
  }
}

\newcommand\orcidicon[1]{\href{https://orcid.org/#1}{\mbox{\scalerel*{
\begin{tikzpicture}[yscale=-1,transform shape]
\pic{orcidlogo};
\end{tikzpicture}
}{|}}}}

\usepackage[nolist,nohyperlinks]{acronym}
\begin{acronym}
\acro{ASMK}{aggregated selective match kernel}
\acro{APE}{absolute positioning error}
\acro{ATE}{absolute trajectory error}

\acro{BA}{bundle adjustment}
\acro{BoW}{Bag-of-Words}
\acro{BRIEF}{binary robust independent elementary features}

\acro{CNN}{convolutional neural network}
\acrodefplural{CNN}[CNNs]{Convolutional neural networks}

\acro{DBoW2}{Bags  of  Binary  Words  for  FAST  Recognition  in  Image  Sequence}
\acro{DOF}{degrees of freedom}

\acro{EKF}{Extended Kalman filter}

\acro{FIM}{Fisher information matrix}

\acrodef{GRU}[GRU]{gated recurrent unit}
\acrodefplural{GRU}{Gated recurrent units}
\acro{GNN}{graph neural network}
\acro{G-CNNs}{Group equivariant Convolutional Neural Networks}

\acro{IMU}{inertial measurement unit}

\acro{KLT}{Kanade-Lucas-Tomasi}

\acro{LIFT}{learned invariant feature transform}
\acro{LSTM}{long short-term memory}
\acrodefplural{LSTM}{long short-term memory networks}

\acro{MSE}{mean square error}
\acro{NCC}{normalized cross correlation}
\acro{NeRF}{neural radiance field}

\acro{ORB}{Oriented FAST and Rotated BRIEF}

\acro{RANSAC}{random sample consensus}
\acro{RCNN}{recurrent convolutional neural network}
\acro{RNN}{recurrent neural network}
\acrodefplural{RNN}[RNNs]{Recurrent neural networks}
\acro{RPE}{relative position error}

\acro{PCA}{principal component analysis}
\acro{PTAM}{parallel tracking and mapping}

\acro{RMSE}{root-mean-square error}

\acro{SAD}{sum of absolute differences}
\acro{SfM}{structure from motion}
\acro{SLAM}{simultaneous localization and mapping}
\acro{SSD}{sum of squared differences}
\acro{SVD}{singular value decomposition}

\acro{UUV}{unmanned underwater vehicle}

\acro{VO}{visual odometry}

\end{acronym}

\begin{document}

%
%
\title{Monocular visual simultaneous localization and mapping: (r)evolution from geometry to deep learning-based pipelines}

\author{Olaya~Álvarez-Tuñón \orcidicon{0000-0003-3581-9481},~\IEEEmembership{Member,~IEEE,}
        Yury~Brodskiy \orcidicon{0009-0002-0445-8126}, 
        and~Erdal~Kayacan \orcidicon{0000-0002-7143-8777},~\IEEEmembership{Senior~Member,~IEEE}

\thanks{This paper is written under the project REMARO which has received funding from the European Union's EU Framework Programme for Research and Innovation Horizon 2020 under Grant Agreement No 956200.
 }

\thanks{ O. Álvarez-Tuñón is with Artificial Intelligence in Robotics Laboratory (AiRLab), the Department of Electrical and Computer Engineering, Aarhus University, 8000 Aarhus C, Denmark  (e-mail: olaya@ece.au.dk).}
\thanks{Y. Brodskiy is with EIVA a/s, 8660 Skanderborg, Denmark (e-mail: ybr@eiva.com).}
\thanks{E. Kayacan is with Automatic Control Group (RAT), Paderborn University, 33098 Paderborn, Germany (e-mail: erdal.kayacan@uni-paderborn.de)}
\thanks{This paragraph will include the Associate Editor who handled your paper.}}

\markboth{Journal of IEEE Transactions on Artificial Intelligence, Vol. 00, No. 0, Month 2020}
{First A. Author \MakeLowercase{\textit{et al.}}: Bare Demo of IEEEtai.cls for IEEE Journals of IEEE Transactions on Artificial Intelligence}

\maketitle

\begin{abstract}

With the rise of deep learning, there is a fundamental change in visual \ac{SLAM} algorithms toward developing different modules trained as end-to-end pipelines.  However, regardless of the implementation domain, visual SLAM’s performance is subject to diverse environmental challenges, such as dynamic elements in outdoor environments, harsh imaging conditions in underwater environments, or blurriness in high-speed setups. These environmental challenges need to be identified to study the real-world viability of SLAM implementations. Motivated by the aforementioned challenges, this paper surveys the current state of visual SLAM algorithms according to the two main frameworks: geometry-based and learning-based SLAM. First, we introduce a general formulation of the SLAM pipeline that includes most of the implementations in the literature. Second, those implementations are classified and surveyed for geometry and learning-based SLAM. After that, environment-specific challenges are formulated to enable experimental evaluation of the resilience of different visual SLAM classes to varying imaging conditions. We address two significant issues in surveying visual SLAM, providing (1) a consistent classification of visual SLAM pipelines and (2) a robust evaluation of their performance under different deployment conditions. Finally, we give our take on future opportunities for visual SLAM implementations.
\end{abstract}

\begin{IEEEImpStatement}
Visual SLAM is one of the fundamental problems in robotics, as it enables autonomous operations in real-world scenarios. Unlike other applications such as object detection or speech recognition, deep learning is yet to be the prevailing standard for visual SLAM. This is partly due to the problem's high dimensionality and the limited data availability.  It is then essential to identify the problem's geometrical nature to determine which learning-based architectures best encode this information. By first introducing conventional approaches for SLAM, this paper surveys prevailing state-of-the-art algorithms while defining the different taxonomies for SLAM and their underlying geometry. Then, deep learning architectures for computer vision are introduced, defining the geometries encoded by their embeddings. This survey critically analyzes the theory behind visual SLAM algorithms. Furthermore, it  evaluates various methods under different environmental conditions, providing insights into the challenges and opportunities for turning deep learning into the new standard for visual SLAM.
\end{IEEEImpStatement}

\section{Introduction}\label{sec:introduction}
\IEEEPARstart{S}{imultaneous} localization and mapping (SLAM) is a well-known method for the autonomous navigation of mobile robots. It defines a state estimation problem in which an autonomous system must determine its location in the environment while generating a representation of it as a map.
\ac{SLAM} has been implemented on autonomous mobile robots using various types of sensors, such as depth cameras, monocular and stereo RGB cameras, LiDAR \cite{intro:survey:lidarkhan2021comparative}, or acoustic \cite{intro:survey:jiang2019surveyacoustic}. This paper focuses on monocular visual-centred approaches for \ac{SLAM}, referred to as monocular visual \ac{SLAM} in the literature.

There are several advantages to using monocular visual SLAM over stereo visual SLAM. One is their simplicity: monocular systems only incorporate a single camera; therefore, they do not require stereo calibration and matching. 
Using a single camera also makes monocular visual SLAM more cost-effective than setups incorporating more sensors. Moreover, unlike stereo systems, the depth of field in monocular systems is not limited by the fixed distance between camera centres, i.e., the baseline. Consequently, monocular visual SLAM systems are more versatile against scene geometry changes than stereo systems.
Nevertheless, the availability of depth information in stereo setups allows them to retrieve the absolute scale of the motion estimate and to provide better accuracy than monocular setups.

\begin{figure*}[ht!]
\centering
\includegraphics[width=.99\textwidth]{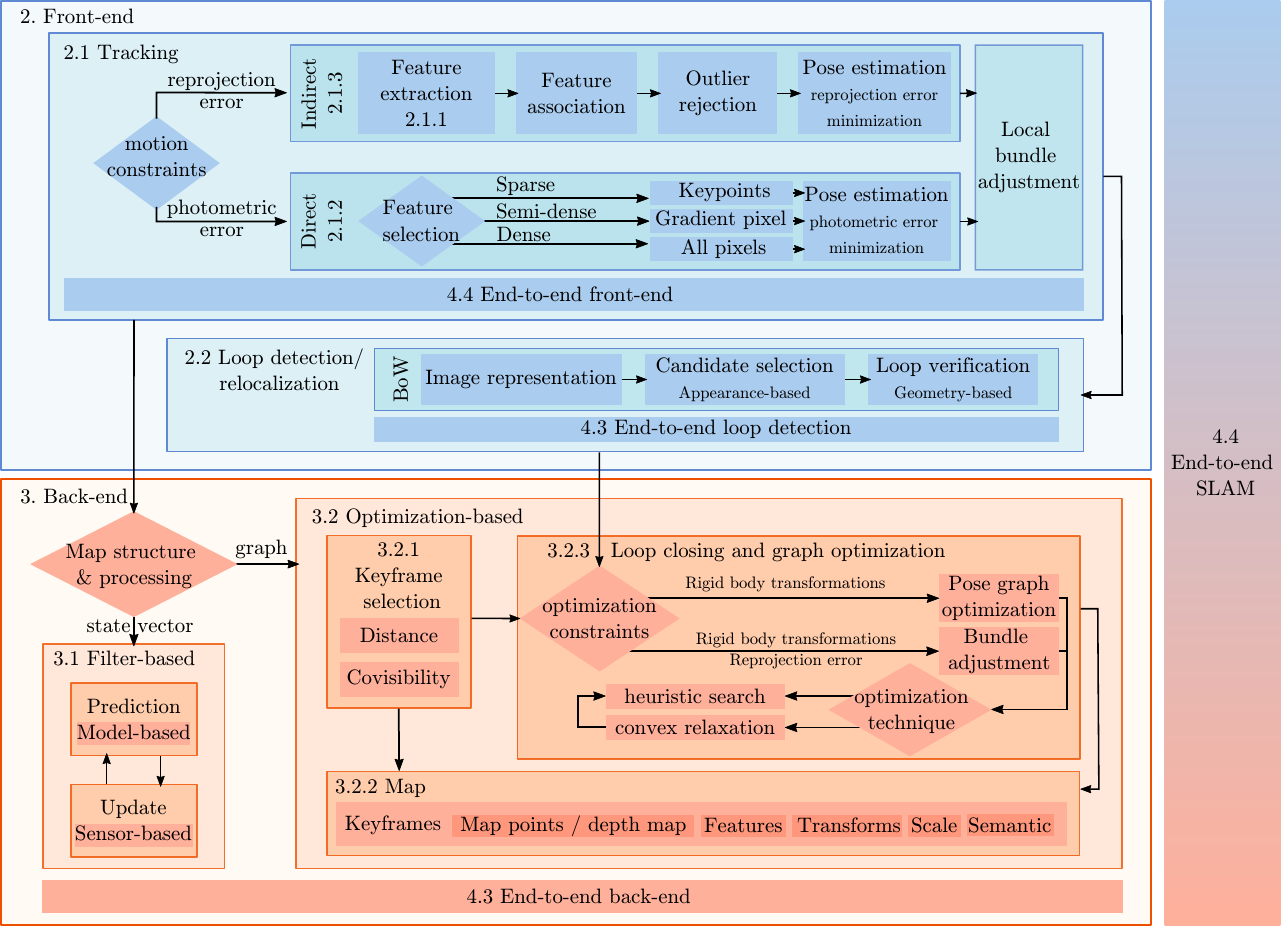}
\caption{Standard modules for SLAM pipeline implementations as outlined in the paper. The front-end is where the raw data is extracted from the sensors and abstracted into a model. The back-end infers from the front-end data and optimizes the estimates. The main modules in the front-end are tracking, loop detection, and relocalization. Tracking can be either direct or indirect. The back-end can be classified into filter and optimization-based techniques.}
\label{fig:pipeline}
\end{figure*}

Early visual \ac{SLAM} applications have relied on filtering algorithms for the estimation task. Given a set of landmarks detected in the image observed by the camera from a particular pose, filtering methods define a joint state vector that includes the landmarks' pose and the camera's aggregated pose. 
A filtering algorithm, e.g. the Kalman filter, iteratively updates this vector. However, as the state vector grows, so does the computational cost, exponentially in some cases \cite{intro:ito2000gaussian}. Moreover, these solutions suffer from random walk noise accumulation over time, which results in drift and high measurement uncertainty \cite{intro:survey:aulinas2008filterslam}.
More recently, a new class of SLAM systems arose: optimization-based or keyframe-based. These systems separate localization and mapping into two parallel threads: the localization, or front-end, is performed at frame rate, while the mapping, or back-end, takes place on a subset of the camera frames (keyframes), at keyframe rate \cite{intro:survey:younes2017keyframe}. 

The front-end includes the estimation of a camera's pose without the computation of a map. This problem is known as visual odometry, and many methods exist to solve it very efficiently. However, as the map is not corrected with new measurements, the pose estimate from visual odometry rapidly drifts over time. 
The back-end handles the map estimate, which in the loop closure algorithm minimizes the front-end drift by iteratively optimizing the map and pose estimates every time an area of the map is revisited. The loop closure algorithm is one of the critical concepts of \ac{SLAM}  that contributes to the camera's precise localization.
The above-mentioned algorithms can be subdivided further, as depicted in Fig. \ref{fig:pipeline}. Visual \ac{SLAM} systems can be further categorized with respect to the selected algorithms.  

Based on the constraints used for the pose estimate in the front-end, a visual \ac{SLAM} system can be classified as indirect or direct.
Indirect \ac{SLAM} techniques minimize the reprojection error between features that need to be extracted first, and associated afterwards. On the other hand, direct \ac{SLAM} methods minimize the photometric error between the selected features. The selection of features can be either sparse, semi-dense, or dense, defining another class of visual \ac{SLAM} systems. 

The back-end structure differs according to whether the approach is filter-based or optimization-based. 
To reduce the state space of the problem, filter-based methods marginalize all past measurements. Optimization-based methods address the state space reduction by performing the measurements under a subset of the frames processed in the front-end, i.e., the keyframes. The keyframes are selected according to distance or covisibility criteria. 
The optimization constraint chosen defines a pose graph optimization or a bundle adjustment algorithm. The information stored by the back-end comprises the \ac{SLAM} map, whose content varies depending on the implementation. It can comprise the 2D features, the 3D (sparse) map points, the depth (dense) map, the relative pose transforms, the estimated scale, and even semantic information.

The SLAM's pipeline has remained unchanged for classic or geometry-based implementations around the above-mentioned configurations. The establishment of deep learning has brought
a fundamental change to that structure, introducing end-to-end implementations of subsets of algorithms within the pipeline, and aiming for end-to-end SLAM approaches.
Visual \ac{SLAM} comprises a multidisciplinary problem, involving computer vision, geometry, optimization, statistics, and deep learning. Except for end-to-end approaches, each SLAM module comprises a set of algorithms implemented independently. The broad scope of the problem leads then to a wide variety of approaches to its study. Previous surveys have reviewed a selection of algorithms developed within a timeline \cite{intro:survey:2010to2016}, or have defined a de-facto standard formulation for SLAM and then reviewed some aspects, such as scalability  and robustness \cite{intro:survey:robust} or time continuity \cite{intro:survey:yan2019flow}. Other surveys focus on specific domain challenges such as dynamic \cite{intro:survey:dynamicenvs} or underwater \cite{intro:survey:uwvisualslam} environments. There are also reviews on subsets of SLAM such as visual odometry \cite{intro:survey:scaramuzza2011visualodomtut,intro:survey:VODL}, loop closure \cite{intro:survey:loopclosure}, pose graph optimization \cite{intro:survey:carlone2015initialization}, map building \cite{intro:survey:geomap, intro:survey:3dmap} or deep learning for pose estimation and semantic information \cite{intro:survey:DLSLAMreview18, intro:survey:semantic20}.

Rather than finding a de-facto implementation for monocular visual SLAM, the present work defines the standard modules that determine its pipeline, and surveys them. To the best of our knowledge, this is the first survey on monocular visual SLAM to introduce a general formulation for its pipeline that involves the different classes of geometry-based SLAM algorithms and end-to-end deep-learning-based SLAM algorithms. The aim of this general formulation is twofold: to serve as an introduction to the SLAM's terminology, and to identify the different combinations of taxonomies that yield a full SLAM pipeline. An outline of the SLAM's modules, as defined in the paper, is shown in Fig. \ref{fig:pipeline}. For each module, the survey studies the recent advances within and outside the context of SLAM. Moreover, the challenges and opportunities for each SLAM algorithm are evaluated throughout the survey, culminating in a final experimental evaluation that compares a set of SLAM classes across different scenarios. 

The outline of the paper is as follows:
The front-end and back-end of geometry-based pipelines are introduced in Sections \ref{sec:frontend} and \ref{sec:backend}. Section \ref{sec:deeplearning} showcases end-to-end pipelines for the different modules for SLAM. Section \ref{sec:environment} enumerates the environmental challenges and shows the results from testing different pipelines under those challenges. Finally, Section \ref{sec:futurework} acknowledges future lines of development for visual SLAM.

\section*{Nomenclature}
\addcontentsline{toc}{section}{Nomenclature}

\begin{description}[\IEEEusemathlabelsep\IEEEsetlabelwidth{$T_{k}, R_k, t_k$}]
\item[$I_k$] image function mapping pixel positions to intensity values (grayscale or RGB).
\item[$C_{k}$] camera pose $\in SE(3)$ in absolute coordinates corresponding to $I_k$. Corresponds to the transform $T_{0,k}$.
\item[$T_{k-1,k}$] rigid body transform $\in SE(3)$ from camera frame $k-1$ to $k$.
\item[$p^k_{i}$] pixel $i$ in image coordinates corresponding to image frame $I_k$.
\item[$\tilde{p}_i$] $p_i$ in homogeneous coordinates.
\item[$\hat{p}_i$] estimated value for $p_i$.
\item[$P_{i}$] projection of $p_{i}$ in 3D coordinates.
\item[$\xi_k$] inverse depth map contained by $I_k$.
\item[$\pi$] function warping the pixels in $I_k$ onto $I_{k-1}$.
\item[$KF$] keyframe.
\end{description}
\section{SLAM: Front-end}
\label{sec:frontend}
Front-end SLAM comprises the lower abstraction layer of SLAM. In the front-end, the raw sensor inputs are abstracted into a model used for tracking and localization.

\subsection{Tracking}
\label{sec:frontend:tracking}
While the overall objective of SLAM is to estimate a globally consistent camera trajectory within a map, the tracking thread incrementally obtains an estimate of the local trajectory.
The tracking module was first implemented as a separate thread in \ac{PTAM} \cite{klein2007ptam}. This way, \ac{PTAM} achieved better performance by removing the need to update the map for each new frame.
The formulation of the tracking thread in monocular visual SLAM is equivalent to that of \ac{VO} \cite{nister2004visual}: it aims to compute the set of relative rigid body transformations $T_{k-1, k}$ between subsequent image frames. 
The global camera poses $C_{0:k}$ can be obtained by concatenating the relative transforms as:
\begin{equation}
    C_{k} = C_{k-1}T_{k-1, k}
\end{equation}

\begin{figure}[b!]

    \centering
    \smallskip
    \includegraphics[width=0.99\columnwidth]{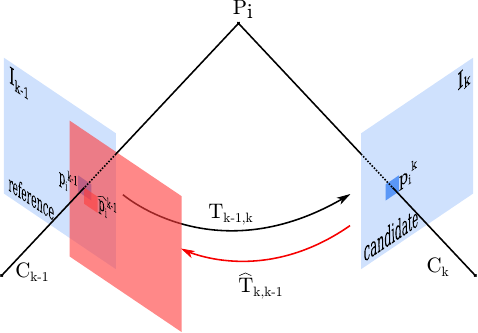}
               \caption{Direct SLAM bases the tracking on optimizing the  photometric error. Given two image frames $I_{k-1}$ and  $I_k$, the 3D point $P_i$ is projected into  $p^{k-1}_i$ and $p^k_{i}$, respectively. The photometric error is the intensity difference between $p^{k-1}_i$ and $p^k_{i}$, obtained by projecting $I_k$ onto $I_{k-1}$.}
               \label{fig:photometric}
\end{figure}

Depending on how motion is estimated from adjacent image frames, we can classify the methods for visual tracking as direct or indirect.
Direct or appearance-based methods estimate the motion from the intensity information of all pixels in the image. Here, the optimization is based on the photometric error, as shown in Fig. \ref{fig:photometric}. Since these methods use more information from the image, they tend to be more robust under areas with low texture and repetitive features, but at a higher computational cost. However, they are more sensitive to illumination changes and large baselines than indirect methods.
Indirect or feature-based methods deduct the motion from the geometric constraints between salient features detected in both images. This computation is based on the optimization of the reprojection error, as shown in Fig. \ref{fig:reprojection}. While computationally inexpensive, these methods discard much valuable information on the image.

In this subsection, direct and indirect methods for tracking in monocular visual SLAM are presented and compared with formulations in the literature.

\subsubsection{Features and descriptors\label{section:features-descriptors}}
Sparse methods use visual features in the images to estimate the camera pose. The neighbourhood around a feature point is called a patch. When a feature is detected, the shape and structure of its patch are embedded into a metric space called descriptor. The descriptors are matched according to a distance function that will depend on the descriptor space: for binary descriptors, it is the Hamming distance, while for histogram-based descriptors, it could be the L1 or the L2 norm. 
Some desired capabilities for a descriptor are invariance under image transformation, equivariance under scale changes, computational efficiency, repeatability and storage efficiency.

The traditional approaches in computer vision have used handcrafted descriptors. Regarding SLAM, the binary descriptors are the primary choice since they achieve good trade-off between robustness and computational efficiency. Some prevailing descriptors for SLAM are \ac{BRIEF}\cite{calonder2011brief}, \ac{ORB}\cite{rublee2011orb}, and A-KAZE \cite{alcantarilla2011AKAZE}.

The  descriptors for \ac{BRIEF} are computed as binary strings from image patches, making them efficient to compute and store \cite{calonder2011brief}. 
Some SLAM implementations using \ac{BRIEF} features are VINS-Mono \cite{qin2018vins-mono}, VPS-SLAM \cite{bavle2020vps-slam} and OV$^2$ SLAM\cite{ferrera2021ov2slam}. VITAMIN-E \cite{Yokozuka_2019vitamine} uses the BRIEF feature without computing the descriptor.
The \ac{ORB} is based on FAST for the features and BRIEF for the descriptors. \ac{ORB} features are computationally efficient and rotation-invariant, and they also achieve robustness in scale by using a scale pyramid of the image \cite{rublee2011orb}. ORB-SLAM \cite{mur2015orb} uses \ac{ORB} features for mapping, tracking, and loop detection.

Handcrafted features are more prone to outliers in the matching than learned features, negatively affecting the tracking precision. Deep learning-based features are richer and thus easier to match, outperforming classical methods in this matter.

State-of-the-art deep learning-based feature extractors  typically compute the features and descriptors simultaneously. For example, \ac{LIFT} \cite{yi2016liftdescriptor}, used in LIFT-SLAM \cite{bruno2021lift}, keeps full end-to-end differentiability when combining feature detection, orientation estimation, and feature description. The training uses 3D points generated by a \ac{SfM} that runs on images at different scales, viewpoints, and lighting conditions.
Superpoint \cite{2018superpoint} operates on full-sized images, rather than just on patches, and is pre-trained with a synthetic dataset first and then automatically adapted to a real unlabeled domain. Unsuperpoint \cite{christiansen2019unsuperpoint} uses a self-supervised approach, making it end-to-end trainable.

\subsubsection{Direct methods}
\label{sec:frontend:directmethods}
Direct methods, introduced in 2007 by DTAM \cite{klein2007dtam}, use image intensities rather than sparse features. They directly compare the pixel intensities between consecutive image frames, so that the pixel values at $p^{k-1}_i$ and at $p^k_{i}$, corresponding to the same 3D point $P_i$ projected in the image planes $I_{k-1}$ and $I_k$, are the most similar to each other. This comparison is defined as a photometric error, which can be formulated as follows:
\begin{equation}
    r_{ph}(P_i,T_{k,k-1},\xi) = \sum_i I_{k-1}(p^{k-1}_i) - I_k\Bigl(\pi (p^{k}_i,T_{k,k-1},\xi)\Bigr)
\end{equation}
with $\pi$ a function warping the pixels in $I_k$ onto the image plane $I_{k-1}$.
Direct methods define tracking as an optimization problem. In this case, the optimization minimizes the aggregated photometric error around a parameter that can either be the camera transform between reference to candidate frame $T_{k-1,k} \in$ SE(3) or the inverse depth map $\xi_k:\Omega\rightarrow \mathbb{R}$, which is directly proportional to the disparity. To each pixel $p^{k}_i$ it corresponds an inverse depth value $d_i = \xi_k(p^{k}_i)$. The use of an inverse depth formulation has the advantage of removing ambiguity, and increasing precision \cite{okutomi1993stereomatch} and, therefore, is more commonly used.

LSD-SLAM \cite{engel2014lsd} implements a Gauss-Newton minimization, iteratively down-weighting large residuals that occlusions and reflections might cause, hence making the tracking more robust against them. GCP-SLAM \cite{zhang2019gcpslam} up-weights the terms with high confidence and down-weights those with high uncertainty.
Some implementations also include the reprojection residual in the optimization \cite{zhang2019gcpslam} \cite{zhao2019robust}. 

The photometric error is a strongly non-convex function; therefore, the minimization can get easily stuck at a local minimum. The two key concepts to consider in addressing this issue are the pixel gradient and the image pyramid.

The pixel gradient defines the intensity change on a pixel. Higher gradients correspond with higher intensity changes and better convergence of the minimization. 
The consideration of the pixel gradient in selecting pixels for minimizing the photometric error yields the following classification of direct methods: sparse, semi-dense and dense.
Sparse direct methods only include sparse keypoints in the computation of the photometric error. In direct SLAM, the descriptors are not computed. 
Semi-dense methods aim to trade off the computational efficiency of sparse methods versus the robustness of dense methods. They reduce the number of pixels tracked by only selecting those with high gradients \cite{forster2014svo}. That is the case for LSD-SLAM \cite{engel2014lsd}, SVO \cite{forster2014svo}, and DSO \cite{engel2017dso}. SVO improves the convexity by computing the photometric error over a patch of pixels rather than a per-pixel computation. 
Dense methods use all pixels in the image, making them more computationally expensive and slower to converge. However, these methods allow computing a complete dense 3D map.

Dense and semi-dense methods are more robust to occlusions or blur due to rapid motion. However, they are more prone to fail under large baselines. Some direct methods handle this issue with an image pyramid, also referred to as coarse-to-fine approach, matching at low resolutions first and then iteratively refining at higher resolutions. Most direct SLAM algorithms follow this approach \cite{klein2007dtam,engel2014lsd,forster2014svo,engel2017dso}. SVO optimizes the photometric error in the coarsest levels until reaching enough convergence and then initialises feature alignment that considers changes in illumination. DSO uses the coarsest pyramid levels to initialise the depth estimate and relocalisation.

\subsubsection{Indirect methods}
\label{sec:frontend:indirectmethods}
The first  feature-based monocular visual SLAM was introduced in MonoSLAM \cite{davison2007monoslam}. Feature-based SLAM tracks sparse salient features between adjacent images.
When tracking sparse features, finding a uniform distribution of the detected features over the image is desired. This way, matching will still be possible in the presence of occlusions, fast movements, or any other situation involving loss of information.
ORB-SLAM \cite{mur2015orb} addresses the search for that uniform distribution by performing a grid search, such that at least five features are detected in each cell. VINS-Mono \cite{qin2018vins-mono} forces a minimum separation of pixels between two neighbouring features.

One standard approach for feature matching would be to detect features, compute their descriptors in two or more views, and then match the closest features in the descriptor space.
These methods are robust to large baselines and view changes. Nevertheless, optical flow methods like the \ac{KLT} tracker are more accurate and therefore better suited for tracking \cite{qin2018vins-mono}. \ac{KLT} approaches match features by minimizing the sum of squared intensity differences \cite{lucas1981klt1}\cite{tomasi1991klt2}.
Another alternative for matching is to assume an affine transformation between adjacent frames and use that transformation to search for the features from the first frame in the second frame. ORB-SLAM's \cite{mur2015orb} pipeline, based on that of \ac{PTAM}  \cite{klein2007ptam}, uses a constant velocity model. It fastens the calculation because the subset of pixels included in it decreases, but it could fail under wrong model estimations or significant imaging changes.
Similarly, VITAMIN-E \cite{Yokozuka_2019vitamine} fits the affine transformation according to the dominant flow estimated from coarse feature pairs. In addition to that, VITAMIN-E performs optimization on the local extrema tracking (rather than descriptors), which avoids falling into a local solution by using the dominant flow estimate.

Classic matching implementations assume static feature points and low or no dynamic feature points. The dynamic feature points are treated then as outliers, which are removed from the matches using outlier rejection algorithms like \ac{RANSAC} \cite{fischler1981RANSACpnp}. Deep learning has introduced semantic methods that identify those dynamic points. This way, they can be removed from the motion computation. Approaches like CubeSLAM \cite{yang2019cubeslam} and PO-SLAM \cite{li2021po-slam} use YOLO \cite{redmon2017yolo9000} to obtain the bounding boxes for dynamic objects. YOL-SLAM \cite{mengcong2021yol-slam} decreases the number of layers in YOLO to improve the processing speed. However, one of the main difficulties for these approaches lies in the ambiguity near the boundaries. The boundaries are typically high-gradient areas, where numerous features might be detected. Methods based on segmentation networks take the boundaries out of the computation.
DS-SLAM\cite{yu2018ds-slam} and SOF-SLAM\cite{cui2019sof-slam} use the segmentation network SegNet \cite{badrinarayanan2017segnet}. DynaSLAM\cite{dynaslam18} uses Mask R-CNN \cite{he2017maskRCNN}, combining it with the geometric consistency for the segmentation. It also fills the occluded background with static information from previous views. RDS-SLAM \cite{liu2021rds-slam} uses BlitzNet \cite{dvornik2017blitznet} for the segmentation mask, with the semantic part as a new thread separated from the tracking, allowing better real-time performance.

Once the features have been matched, the motion can be estimated. This extra step of estimating the motion from the geometry constraints, rather than directly estimating it from the image intensities, gives this family of methods the qualification of "indirect".
 For monocular visual SLAM, there are two cases for motion extraction: 2D-to-2D, and perspective-n-point (PnP), also referred to as 3D-to-2D.
 
In {2D-to-2D}, and assuming calibrated cameras, the essential matrix can be derived from the matches using the epipolar constraint. It follows that, for the matched points $p^{k-1}_i \leftrightarrow p^{k}_i$,    \begin{equation}
        \left[ \tilde{p}^{k}_i \right]^T E \tilde{p}^{k-1}_i = 0
    \end{equation}
    in homogeneous coordinates. The essential matrix $E$ has five degrees of freedom: It can be solved with at least 5 points \cite{nister2004-5point}, although the most classic implementation is the eight-point-algorithm \cite{hartley1997-8point}.
    Rotation and translation can be derived from it up to an scale factor.
    \begin{equation}
        E_k = [t_k]_\times R_k, 
    \end{equation}
    with $[t]_\times$ the matrix representation of the cross product of $t$. $R_k$ and $t_k$ are obtained from the \ac{SVD} of $E$. For further details on the theory behind epipolar geometry, we refer the reader to the book by Hartley and Zisserman \cite{heyden2005multiplevgHyZ}. This method suffers from instability under small translations and is unsolvable under pure rotations, thus is not usually implemented in the front-end. Instead, it has been used for map initialization \cite{butt2020epipolarinitializ} or map point verification \cite{mur2015orb}. 
PnP is the problem of estimating the relative pose between a calibrated camera and $n$ 3D points in the scene, given the correspondences of their 2D projections \cite{fischler1981RANSACpnp}. It requires matched features across three views, which requires an initialization in the monocular case. The minimal solution is given by the Perspective-three-point (P3P) algorithm \cite{gao2003p3p}. ORB-SLAM \cite{mur2015orb} (and by extension, all approaches using the ORB-SLAM front-end) uses EPnP, which can use any amount of points \cite{lepetit2009epnp}. DF-VO \cite{zhan2019dfvo} uses a learnt optical flow for the 2D correspondences and a learnt depth for the 3D representation.

\begin{figure}[b!]

    \centering
    \smallskip
    \includegraphics[width=0.99\columnwidth]{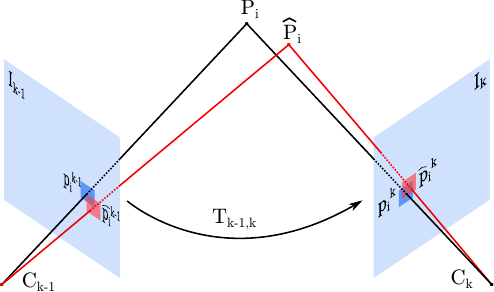}

    \caption{Sparse SLAM optimizing the reprojection error for tracking. Two matched points $p^{k-1}_i$ and $p^{k}_i$ are triangulated onto the estimated 3D point $\hat{P}_i$, which is then reprojected as $\hat{p}^{k-1}_i$ and $\hat{p}^{k}_i$. The error is the difference between reprojected and original position of the points.}
    \label{fig:reprojection}
\end{figure}

Once the first motion estimate is obtained, the measurements are refined in an optimization step, often referred to as local bundle adjustment. 
It consists of minimizing the reprojection error, resulting from projecting the 3D points into the image planes and comparing the reprojection with the original position of the feature points, as represented in Fig. \ref{fig:reprojection}. 
The reprojection error for the matched points $\textbf{p}^{k-1}_i \leftrightarrow \textbf{p}^{k}_i$, can be formulated as a least-squares problem:
\begin{equation}
     r_{r}(P_i) = \sum_id(p^{k-1}_i,\hat{p}^{k-1}_i)^2+d(p^{k}_i,\hat{p}^{k}_i)^2,
\end{equation}
with $d$ the distance function.

\subsection{Loop closure detection and relocalization}
\label{section:frontend:LoopClosure-relocalization}
The loop closure thread aims to find a revisited area in the front-end and then optimize the map and trajectory in the back end.
Similarly, if the robot loses track of its trajectory, it can relocate itself in the previously built map by detecting a loop. Consequently, loop detection is one of the essential modules for the accuracy and robustness of \ac{SLAM}.

\begin{figure}[b!]
  \centering
  \includegraphics[width=.9\columnwidth]{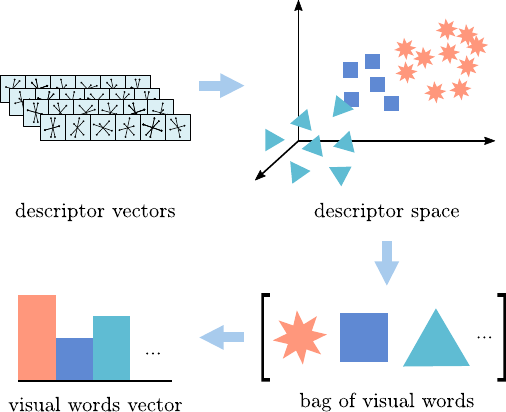}
  \caption{The BoW framework represents the image by extracting
  and clustering the descriptor vectors. Each cluster represents a visual word. The histogram of words comprises the BoW vector that describes the image's appearance.}
\end{figure}

The key to loop closure detection lies in calculating the similarity between images. An intuitive approach would be determining a loop according to the number of correct matches detected in feature matching. However, searching through the whole database will be computationally expensive and memory inefficient as the map grows.
Instead, loop detection is divided into three steps: image representation, candidate selection, and loop verification.

One classical framework for loop detection is \ac{BoW}, first proposed in \cite{sivic2003bowzisserman}. In this framework, image features and their descriptors are extracted for the image representation. ORB-SLAM \cite{campos2021orb} uses DBoW2\cite{galvez2012dbow2}  with the same ORB features as in tracking to achieve real-time performance. Other approaches leveraging the ORB-SLAM backend also use DBoW2, namely LIFT-SLAM \cite{bruno2021lift}, RDS-SLAM \cite{liu2021rds-slam}, SOF-SLAM \cite{cui2019sof-slam}, and DS-SLAM \cite{yu2018ds-slam}. Deep learning features also find their application in loop closure detection, not only with patch descriptors \cite{LCDwithSuperpoint}, but also with global descriptors \cite{li2020dxslam,xu2021esa-vlad}.
The descriptors are then clustered and stored in a codebook with a clustering algorithm like k-means. Each cluster represents a visual word of the codebook. Additionally, DBoW2 \cite{galvez2012dbow2} repeats this operation for each cluster to generate a hierarchical tree. The clustering of descriptors is time-consuming and usually trained offline. The work in iBoW-LCD \cite{garcia2018ibow}, implemented in 
OV$^2$SLAM \cite{ferrera2021ov2slam}, builds the vocabulary trees incrementally so that it constantly adapts to the current environment without needing offline training.

The candidate frame selection retrieves candidates according to the nearest neighbours to the queried frame. 
Most implementations rely on image sequences rather than single images.
DBoW2 \cite{galvez2012dbow2} chooses consistent candidates within a time interval before entering the verification step.

BoW portrays an appearance-based framework and does not encode spatial information. Therefore, loop verification applies a geometry consistency check. It is done by finding a fundamental matrix in DBoW2 \cite{galvez2012dbow2} and also in ESA-VLAD \cite{xu2021esa-vlad} between the queried and candidate images with enough inliers using RANSAC. ORB-SLAM2 \cite{campos2021orb} additionally applies place recognition to three consecutive frames to improve recall. DXSLAM \cite{li2020dxslam} and OV$^2$SLAM \cite{ferrera2021ov2slam} apply the standard RANSAC and PnP process.
In LRO \cite{ma2021loopsuperpoint2} the loop is verified according to the topological structure between groups of keypoints, considering rotation, scaling, and deformation.

Computationally expensive SLAM approaches sometimes sacrifice the loop closure thread to speed up the algorithm. That is the case of direct slam approaches \cite{klein2007dtam,zhang2020vdo,li2021po-slam}, or in those that incorporate deep learning in some of their processes \cite{tateno2017cnnslam,dynaslam18,fan2022blitz-slam}. However, some deep learning approaches utilize their capabilities in the loop closure thread. That is the case of some semantic approaches that detect the loop when reobserving an object in the scene \cite{bavle2020vps-slam,parkhiya2018semanticloopclosure}.

Direct approaches like LSD-SLAM \cite{engel2014lsd}, the subsequent DSO \cite{engel2017dso} and GCP-SLAM \cite{zhang2019gcpslam} use the appearance-based method FAB-MAP \cite{cummins2008fabmap} for detecting a loop, which is verified with a statistical test.  
The work in LDSO \cite{gao2018ldso} extends the capabilities of DSO by using the classic BoW framework and the same features as in the front-end.

\section{SLAM: Back-end}
\label{sec:backend}
With large-scale environments, the need for a global map arises as the front-end estimate tends to drift over time.
Back-end infers in the front-end data, abstracting it into a map and optimizing it. According to how the map is structured and processed, there are two approaches: filter-based and optimization-based.

Filtering approaches marginalise all the poses for every new frame while retaining all the landmarks detected, creating a compact graph. This graph only grows when exploring new areas and detecting new features. However, it will eventually become a fully interconnected graph, severely limiting the number of features that can be stored while preserving computational efficiency. 

\begin{figure}[!b]
 \centering

\subfloat[]{\includegraphics[width=0.47\linewidth]{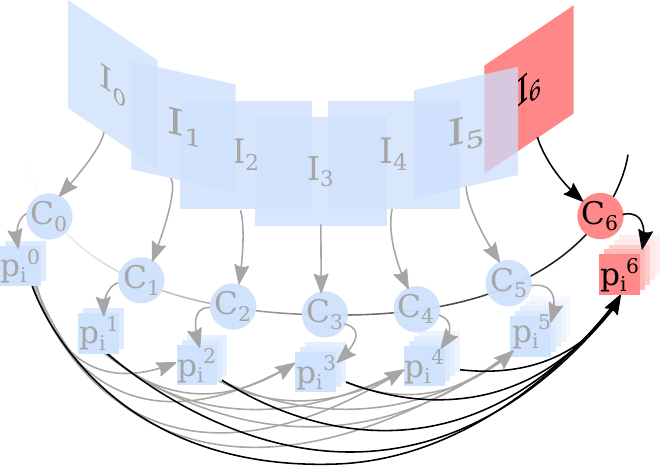}}\quad
\subfloat[]{\includegraphics[width=0.47\linewidth]{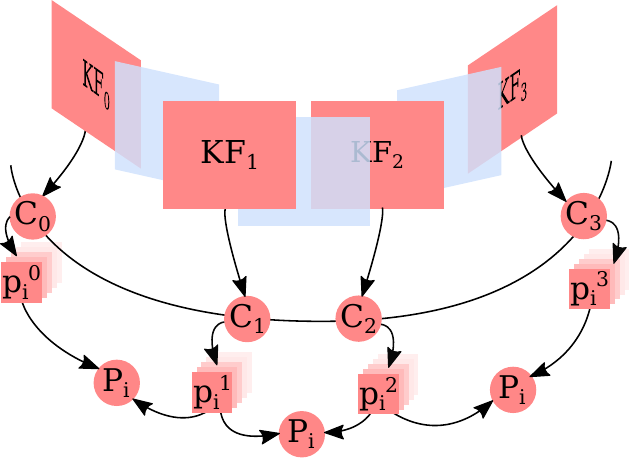}}

\caption{According to its back-end, SLAM can be classified as (\textbf{a}) filter-based, and  (\textbf{b}) keyframe-based. Filter-based approaches marginalise all camera poses onto the last pose $C_k$, which is stored in the state vector with all landmarks $p_k$. Keyframe-based approaches only process data from the keyframes $K_i$, generating a sparse graph with the camera poses $C_k$, the landmarks $p_k$, and the map points $P_i$.}
\centering
\label{fig:filtervsopt}
\end{figure}

Optimization approaches, as opposed to filtering, solve the complete graph, leveraging an algorithm such as \ac{BA}. They can either do it in a sliding window (local \ac{BA}), or in a subset of the overall frames, referred to as keyframes (global \ac{BA}). The rate at which the optimization is performed is lower than the frame rate: despite including more elements in the computation, they are sparsely connected, making optimization approaches comparatively more efficient.
Furthermore, it has been argued that adding features is more beneficial for accuracy than adding frames \cite{strasdat2010whyfilteristhequestion}. All these facts have turned optimization approaches into the new de facto standard for the back-end in SLAM. A comparison between filtering and optimization approaches is represented in Fig. \ref{fig:filtervsopt}.

In this section, filter-based approaches will be briefly outlined, followed by the introduction of the relevant elements and procedures for optimization-based approaches.

\subsection{Filter-based}
\label{sec:backend:filterbased}

Filter-based approaches present SLAM as a probabilistic framework formulated with variants of the Bayes filter. These frameworks represent the environment and the robot's variables (i.e., pose and velocity) as a continuous state vector. The state is modelled as a Markov chain, in which the current belief contains all the knowledge about the previous states of the robot, that is, the probability of the next state depends only on the current one. This property originated from Markov's assumptions of a perfectly modelled system, a static world, and noise independence over time. While these assumptions do not hold in the real world, the uncertainty modelling from probabilistic  approaches provides robustness against their violations \cite{probroboticsthrun}. 

In filter-based SLAM, the state vector is updated recursively according to its belief distribution, based on the available data. This update is performed in two steps: the prediction step, based on the previous state posterior, and the update or correction step, which incorporates the measurements originating from the environment's state. However, the Bayes filter framework only considers linear distributions under additive Gaussian noise.

\ac{EKF} SLAM is one of the first approaches to state estimation for SLAM \cite{probroboticsthrun}, and it addresses the system's nonlinearities by linearizing the posterior around the state while keeping the Gaussian noise assumption.  
It was first implemented in a monocular visual SLAM system in MonoSLAM \cite{davison2007monoslam}. EFK SLAM represents the belief for the state vector by its first and second moments. In MonoSLAM, the state vector contains the 3D pose of the camera plus the 3D position vector for the features. It formulates the prediction step according to a constant linear and angular velocity and the update step according to the measurements of the image features. Some more recent implementations of Kalman filters for SLAM include ROVIO \cite{bloesch2015rovio}, and S-MSCKF \cite{sun2018s-msckf}, which also couple inertial measurements into the state vector and the process model.
    
Particle filters can work for nonlinear systems and non-Gaussian noise distributions \cite{pupilli2005realtctpf}. On the other hand, the main limitation of particle filters is the large number of particles required for high-dimensional spaces. Some SLAM systems overcome this issue by implementing Rao-Blackwellized Particle Filters (RBPF) for maintaining the distribution of the poses and the features in the map \cite{huletski2015modernevaluation}. L-SLAM \cite{zikos20156L-SLAM} applies the particle filter to the parameter that induces the nonlinearities, that is, the orientation, and the Kalman filter to the parameters that can be represented in a linear form, that is, the translation along with the position of the features.

Although filter-based approaches still find their applications in SLAM \cite{geneva2019SEVIS,quan2019accurate,heo2018ekf,geneva2019linear}, optimization-based techniques have taken over the lead in visual SLAM implementations and therefore will be the focus of the rest of the sections of the back-end survey.

\subsection{Optimization-based}
\label{sec:backend:optbased}
Most recent methods are based on optimization approaches, which perform local camera tracking in the front-end and optimize the map and the camera trajectory in the back-end. Both processes run on parallel threads. 
Optimization-based SLAM relies on keyframes to reduce the computational cost of processing consecutive frames. Every time a new keyframe comes in, an optimization procedure is applied over the comprehensive set of keyframes, througout the map and trajectory estimates.
Moreover, the optimization is also performed when a loop closure is detected in the front-end, allowing detection and correction of the accumulated drift.
The information stored by the map and its structure will depend not only on the information processed by the front-end but also on the optimization process and the desired output data from the overall SLAM algorithm.

The current section introduces the fundamental concepts relevant to the back-end for optimization approaches of SLAM: the criteria for keyframe selection, the information stored in SLAM maps, and lastly, the loop closing and optimization of the SLAM graph.

\subsubsection{Keyframe selection}
\label{sec:backend:keyframeselection}
The back-end runs on the information extracted from the keyframes; hence it is crucial to define optimal criteria for selecting whether the frame retrieved by the front-end is a keyframe.
PTAM \cite{klein2007ptam}, LSD-SLAM \cite{engel2014lsd}, and DTAM \cite{klein2007dtam} include keyframes according to a distance criterion. VINS-Mono \cite{qin2018vins-mono} applies a parallax criterion, which considers translation and rotation. Moreover, VINS-Mono performs the keyframe selection in the front-end, previous to tracking.
ORB-SLAM \cite{campos2021orb} is generous when including keyframes but very restrictive for keeping them, removing those that include redundant map points. ORB-SLAM and SVO \cite{forster2014svo} keep them according to visual change criteria as a function of the keypoints tracked. VOLDOR+SLAM \cite{min2021voldor+} uses a covisibility criteria of the map. OV$^2$SLAM \cite{ferrera2021ov2slam} uses both a parallax and keypoint tracking criteria, with a posterior culling similar to that of ORB-SLAM.

\subsubsection{Information stored in the map}
\label{sec:backend:map}
The map contains information about the environment intended to be used by other threads within the SLAM pipeline, namely loop closure or third-party processes such as augmented reality or path planning algorithms.

\begin{figure}[!b]
 \centering

 \subfloat[]{\includegraphics[width=\columnwidth]{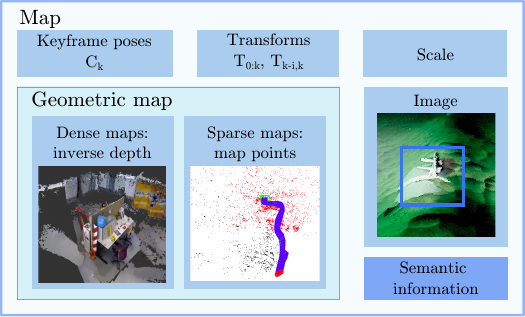}}\quad
\subfloat[]{\includegraphics[width=\columnwidth]{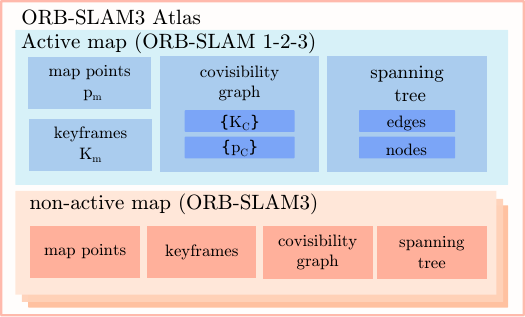}}

\caption{SLAM maps comprise the information inferred from sensor's measurements. \textbf{(a)} showcases the elements that a SLAM map can contain.  \textbf{(b)} shows the structure of the Atlas map from ORB-SLAM1, 2, and 3. The notation for the map elements is the same as in the original papers.}
\label{fig:atlasmap}
\centering
\end{figure}

The information stored by the map, as obtained from the keyframes, is summarized in Fig. \ref{fig:atlasmap}a.
A minimal setup includes the camera poses and the transforms between keyframes and the world coordinates. 
Some maps include the image from the camera input. Additionally, approaches like PTAM \cite{klein2007ptam} store a pyramid of subsampled grayscale images. However, other approaches discard this data to reduce memory consumption, like VINS-Mono \cite{qin2018vins-mono}.

The geometric map could be sparse or dense. Generally, sparse features are meant explicitly for localization, while dense features also allow scene representation. Sparse maps comprise the triangulated 3D points in world coordinates, with reference to their pixel location in the corresponding pyramid level of the keyframe. To aim for robust map points, approaches like ORB-SLAM \cite{campos2021orb}, or UcoSLAM \cite{munoz2020ucoslam} apply survival strategies consisting in keeping only those that are visible in a window of keyframes.
Dense or semi-dense maps like the one generated by LSD-SLAM \cite{engel2014lsd} store a semi-dense inverse depth map for the pixels with larger gradients. Although usually direct methods are the ones that generate depth maps \cite{engel2014lsd,tateno2017cnnslam,czarnowski2020deepfactors}, indirect methods are also able to generate depth maps from sparse points: using a high amount of points in VITAMIN-E  \cite{Yokozuka_2019vitamine}, detecting planar surfaces from semantic segmentation in VPS-SLAM \cite{bavle2020vps-slam}, filling the background in DynaSLAM \cite{dynaslam18}, or creating a dense semantic octo-tree map in DS-SLAM \cite{yu2018ds-slam}. 
As an alternative to classic dense or sparse feature representations, the work in \cite{zhen2021mapquadrics} proposes using simplified quadrics for geometric representation. It allows the creation of compact maps while still capturing layout.

Semantic information extends the capabilities of SLAM by adding information that other processes can use running in the system, most of them based in deep learning methods \cite{tateno2017cnnslam,liu2021rds-slam,yang2019cubeslam,bavle2020vps-slam,cui2019sof-slam,dynaslam18}. Semantic approaches are computationally expensive and cannot be implemented in real time. DS-SLAM \cite{yu2018ds-slam} proposes creating the semantic map  in a separate thread to improve the real time performance.

The inherent scale ambiguity of monocular systems makes SLAM drift. In the absence of other sensors to fuse the measurement with, classic approaches explicitly detect the drift in loop closures by normalizing the inverse depth map \cite{engel2014lsd}\cite{campos2021orb}. On the other hand, deep learning methods like CNN-SLAM \cite{tateno2017cnnslam} or UnDeepVO \cite{li2018undeepvo} can recover the absolute scale from monocular images\cite{czarnowski2020deepfactors} \cite{greene2020metrically}. They minimize the drift by choosing an adequate loss function for the network, for instance, by exploiting epipolar constraints \cite{godard2017unsupervisedleftrightconsistency} or the scale consistency between frames \cite{bian2019unsuperviseddepthconsistency}.
VDO-SLAM \cite{zhang2020vdo} estimates the depth map with the learning-based monocular depth estimation method MonoDepth2 \cite{godard2019monodepth2}.

Designing a mapping process implies meeting a tradeoff between tracking robustness and memory efficiency.
The first versions of ORB-SLAM were susceptible to tracking loss: if the algorithm could not track the pose in the local map, the tracking loss would require restarting the process. However, since ORB-SLAM3 \cite{campos2021orb}, this issue is addressed by the so-called Atlas map (see Fig. \ref{fig:atlasmap}b). The Atlas map stores as inactive those maps in which the tracking was lost and initializes a new map from that frame on.  
These maps can be later merged if the active map revisits an area from an inactive map. 
Multi-map strategies improve relocalization under tracking loss or long mapping sessions and allow collaborative mapping.

\subsubsection{Loop closing and graph optimization}
\label{sec:backend:loopclosingandgraphopt}

Introducing keyframes has allowed the creation of larger maps and more computationally expensive but accurate optimization algorithms for SLAM. Optimization methods for SLAM are based on factor graphs. In general, when applied to the \ac{SLAM} problem, whereas the keyframes represent the nodes, and the edges represent the relationship between them, specific to each implementation.
\begin{figure}[!b]

    \centering
    \smallskip
    \includegraphics[width=0.9\linewidth]{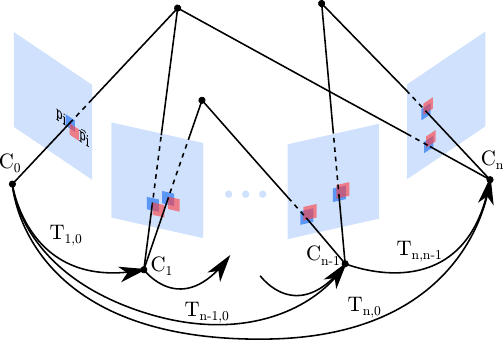}

    \caption{Monocular visual SLAM uses factor graphs to optimize the data gathered in the front-end. In this figure, each image frame represents in red the observed value $\hat{p}_{ij}$ for the 3D point $P_i$ , and in blue its reprojection value $p_{ij}$. Bundle adjusment optimizes the camera poses, the 3D map points and their respective 2D projections. Pose graph optimization optimizes the camera poses and the rigid-body transformations.}
\end{figure}
There are two main implementations of factor graphs for SLAM: pose graph optimization and  bundle adjustment (BA).

In pose graph optimization the nodes are the camera poses $C_k$, and the edges the rigid-body transformations $T_{jk}$. The optimization problem that finds the optimal camera poses can be formulated as:    
\begin{equation}
        \argmin{C_k,T_k} \sum_{k,j} \rho(r(C_k,C_jT_{jk})) 
    \end{equation}
     with $r$ the residual error based in a distance function, and $\rho$ the cost function for the residual error. It is used by frameworks that require a lower computational cost in the back end, such as LSD-SLAM \cite{engel2014lsd} and CNN-SLAM \cite{tateno2017cnnslam}.

\ac{BA} also stores the position of the map points, in addition to the information stored by the pose graph. The edges are now the reprojection constraints from the points to the cameras. Given $N$ 3D points $P_i$ seen from $M$ camera poses $C_k$, with $p_{ki}$ the observed 2D projection of the point  $P_i$ in the camera $C_k$, projected by the projection matrix $\pi_k$, the optimization problem for the reprojection error can be formulated as \cite{heyden2005multiplevgHyZ}:
    \begin{equation}
        \argmin{\hat{p}_{ki},P_i,\pi_k} \sum^N_i \sum^M_k \rho(r(\hat{p}_{ki}, \pi_kP_i))
    \end{equation}
     In local \ac{BA}, the optimization is performed over a window of connected keyframes, leaving the rest fixed, while global \ac{BA} operates over all keyframes within the map. Since it uses more data for optimization, \ac{BA} is more precise and computationally costly. PTAM applies local \ac{BA} for each newly inserted keyframe, and then global \ac{BA} in a separate thread. For dense point clouds, the computational cost of \ac{BA} is unfeasible; thus, VITAMIN-E  \cite{Yokozuka_2019vitamine} proposes an method called subspace Gauss-Newton. Instead of updating all the variables at once, it partially updates a subspace of them, keeping the rest fixed. The semantic approach CubeSLAM \cite{yang2019cubeslam} takes into account which points are static and which are dynamic and applies motion model constraints to the dynamic points.

Optimization-based SLAM algorithms have commonly defined a nonlinear least-squares problem, in which the cost function $\rho(\cdot)$ corresponds to the square of the residual.

Some approaches run combinations of the methods mentioned above. A common approach is running a local \ac{BA}, then a pose graph optimization for loop closing \cite{qin2018vins-mono,bruno2021lift,cui2019sof-slam,mur2015orb,liu2021rds-slam}.
That was the case for the first version of ORB-SLAM \cite{mur2015orb}. Since ORB-SLAM2 \cite{mur2017orbslam2}, a new thread running full (or global) \ac{BA} was introduced. The same ORB features are used for better real-time performance for all these processes. As an alternative to full \ac{BA}, OV$^2$SLAM decreases the runtime by applying the so-called "loose-BA", which only optimizes the keyframes affected by the loop closure. 

\begin{figure}[!b]

    \centering
    \smallskip
    \includegraphics[width=1\columnwidth]{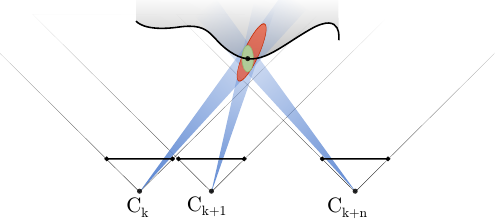}

    \caption{When observing a patch corresponding to the same 3D point from different baselines (for simplicity, depicted as a pure linear displacement of the camera), lower baselines result in higher uncertainty (in red), while higher baselines result in lower uncertainty (in green) for inverse depth estimation.}
    \label{fig:disparity}
\end{figure}

Some limitations of the \ac{BA} process arise under the presence of outliers, pure rotational motions, and small baselines. The nonlinear least-squares optimization process presents a strong sensitivity to outliers, which can dominate the cost function and lead to wrong results.
With pure rotational motion or without enough baseline, the depth uncertainty grows due to unobservable disparities and numerical errors \cite{hess2014uncertaintyandbaseline}, as represented in Fig. \ref{fig:disparity}. The work in \cite{bustos2019LinfinitySLAM} proposes rotation averaging \cite{eriksson2018rotationaveraging} during the online update of the graph, leaving the global graph optimization as a lower-priority thread. This pipeline shows more robustness under pure rotational or low-speed motion updates while simplifying the online computation.
Moreover, the least-squares approach for the cost function is sensitive to outliers from poor initialization, bad front-end estimates, or wrong loop estimates. This is due to the quadratic cost function making the measurements with high residuals have a big influence on the overall cost. Robust estimation theory suggests using robust cost functions such as the truncated least-squares, which prevents large residuals from dominating the cost, or the Huber loss, which is convex \cite{black1996robustloss}. However, the sum of all cost functions remains non-convex. 

The aforementioned methods are based on a heuristic search, which, while computationally efficient, are vulnerable to convergence to local minima. Alternatively, convex relaxation techniques search for convex approximations that provide better convergence.
The work in \cite{yangcarlone2020GNC} proposes a method called graduated non-convexity (GNC), which iteratively solves the optimization problem. It starts from a convex surrogate of the loss and gradually increases the non-convexity until the original non-convex least-squares formulation is reached. Despite being more computationally expensive, this method proved a better performance than classic solvers under the high presence of outliers.
On the other hand, SE-Sync \cite{backend:sesyncrosen2019se} defines a simplified maximum likelihood estimation of the Euclidean synchronization problem for pose graph optimization. A convex semidefinite relaxation is applied to the estimator, which is then reduced to a low-dimension Riemannian manifold. Finally, the optimally-global solution is found with a Riemannian optimization method. Moreover, SE-Sync provides assessment on the global optimality of the found minimization.

Although the methods based in heuristic search such as Gauss-Newton and Levenberg-Marquard are more established within visual SLAM pipelines, convex relaxation techniques present a promising alternative to be seen in future implementations.
\section{Deep learning: towards end-to-end SLAM}
\label{sec:deeplearning}

End-to-end monocular visual SLAM models are as diverse as deep learning networks for computer vision. Understanding how the networks encode spatial, temporal, contextual, and semantic information is necessary to comprehend how each architecture influences the expected behaviour of the models. A camera moving within a euclidean space SE(3) involves the space group symmetries; translation and rotation. Moreover, projecting the euclidean space onto an image plane defines a projective space in which scale symmetry must also be considered. Choosing the correct architectural elements to appropriately encode symmetries within a neural network is essential for successfully creating complex learned systems such as monocular visual SLAM.

Several deep learning architectures have arisen in the last decade to explicitly support learning from spatial data \cite{surveyDLspatialdata}.
Convolutional layers, also referred to as Euclidean or planar convolutions, are shift-equivariant; a shift to the input produces a shift to the output feature maps \cite{bronstein2021geometricDL}. Pooling layers allow scale invariance by locally subsampling the input feature map. 
\Acp{CNN} in object detection achieve scale and rotation equivariance by feeding rotated and scaled versions of the training data to the network. However, localization pipelines require more generalized underlying representations. 
\ac{G-CNNs} exploit the shift-invariance of Euclidean convolutions  by rotating and transforming them, thus achieving rotation and scale equivariance in discrete groups \cite{dl:groupconv}. Spherical \acp{CNN} \cite{dl:sphericalCNN} achieved equivariance within the continuous group for rotations SO(3).

Aside from spatial information, visual localization shows a strong temporal correlation between visual features in consecutive frames.  
\Acp{RNN} carry out temporal representations by sequentially encoding data. \acfp{GRU} and \acp{LSTM} encode the temporal information with the so-called gating mechanisms, which regulate the signal flow across the network.
If implementing an appropriate range of gate values, the gating mechanisms achieve invariance to time transformations \cite{dl:rnn:canwarptime}. However, the performance of \acp{RNN} is hindered by their inability to parallelize under longer sequences.
Transformers overcome the sequential processing of data by implementing the so-called attention mechanisms \cite{dl:attentioniswhatyouneed}, which allow for parallelization. They are originally permutation-invariant \cite{bronstein2021geometricDL}, but have been recently extended to provide equivariance within Lie groups \cite{dl:lietransformer}. Recently, \acp{GNN} have gained popularity, since many real-world problems represent data as graphs \cite{dl:intro:xia2021graphsurvey}. Initially designed as a permutation invariant or equivariant architecture, \acp{GNN} encode the geometric structure and local invariance of the data \cite{keriven2019gnninvoreq, dl:intro:jiao2022graphvisionsurvey}. \acp{GNN} benefit from the computational efficiency of sparse representations \cite{dl:intro:sparselearning}. Moreover, they allow for implementations of the mechanisms mentioned above, such as attention \cite{velivckovic2017graphattention}, convolution \cite{welling2016convgnn} and pooling \cite{yuan2020structpool}. 

The current section surveys end-to-end pipelines for each SLAM module with a focus on the used architectural solutions.

\subsection{End-to-end loop detection}
\label{sec:deeplearning:loop}

End-to-end loop detection pipelines comprise appearance-based methods that require robustness to lighting changes and different viewpoints. The baseline architectures for image representation rely on \ac{CNN}-based descriptors, where lower layers represent low-level features such as corners or edges, and upper layers represent semantic features such as buildings or faces. 

\begin{figure}[b!]
  \centering
  \includegraphics[width=.9\columnwidth]{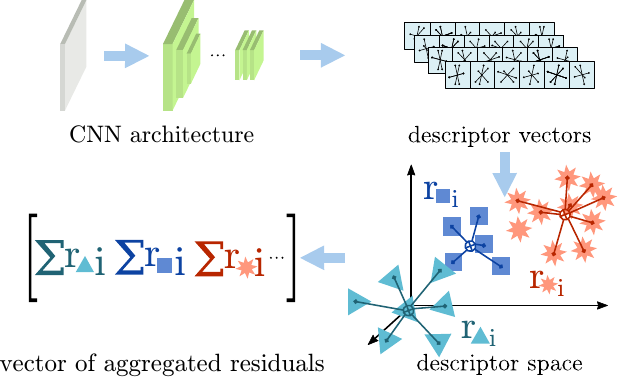}
  \caption{ General framework of the end-to-end methods for loop detection. The descriptors are obtained from pre-trained classification networks. Then, they are clustered and assigned a residual according to the centroid of the cluster they belong to. For each cluster, the residuals are aggregated into a vector that describes the image's appearance.}
  \label{fig:end2endloopdetection}
\end{figure}

One of the first end-to-end loop detection architectures was proposed in NetVLAD \cite{arandjelovic2016netvlad}. The dense description implements the \ac{CNN} architectures from AlexNet \cite{krizhevsky2012imagenet-alexnet} and VGG-16 \cite{dl:loopclosure:vgg16} as the baseline, cropped at the last convolutional layer. Rather than using the \ac{BoW} approach of concatenating clusters of descriptors, their residuals are aggregated into a single vector, for better memory efficiency (see Fig. \ref{fig:end2endloopdetection}). Furthermore, \cite{zhu2018attention,arandjelovic2016netvlad} apply PCA whitening to the descriptor, scaling its dimensions to address the issue of over-counting and co-occurrences.

Deep learning models aim to provide invariance to scale, translation, rotation, and illumination changes.
One method for achieving scale invariance is the implementation of spatial and global pooling layers \cite{zhu2018attention,ento-to-endloop,loopnet-attention}. In addition to that, other implementations use multi-scale descriptors \cite{arandjelovic2016netvlad,chen2018learningPR}, that is, they extract features at different layers in the network. A slightly different approach is followed in LoopNet \cite{loopnet-attention}, which inputs scaled images at different stages of the network.

Instead of using standard metrics for measuring the distance between descriptors, the end-to-end descriptors in \cite{ento-to-endloop,arandjelovic2016netvlad} apply metric learning. In other words, they learn a loss that provides a smaller distance between the query and the best candidate than between the query and a set of negative candidates.

The introduction of attention mechanisms allows for models to focus in those regions on the image that are more relevant to the localization task. This way, the model can suppress the scene's transient information. 
They can identify visually-salient areas in the image \cite{zhu2018attention}, or also those of semantic importance \cite{chen2018learningPR}, and weight the descriptors accordingly prior to their aggregation. 

The aforementioned methods comprise appearance-based image retrieval algorithms. With NetVLAD \cite{dl:loopclosure:relposegnn} or a pre-trained CNN \cite{dl:loop:li2021transcamp,dl:loopclosure:li2022gtcar} as baseline, leveraging \acp{GNN} has allowed to directly regress the absolute pose of the candidate frames. 

\subsection{End-to-end front-end}
\label{sec:deeplearning:frontend}
The previous section introduced end-to-end approaches for loop detection. Loop detection defines an appearance-based localization task, which results in image retrieval of the closest match to a query image. However, the context of monocular visual \ac{SLAM} requires to retrieve the camera pose for the queried image, either for tracking or relocalization. 

\begin{figure}[b!]
  \centering
  \includegraphics[width=.95\columnwidth]{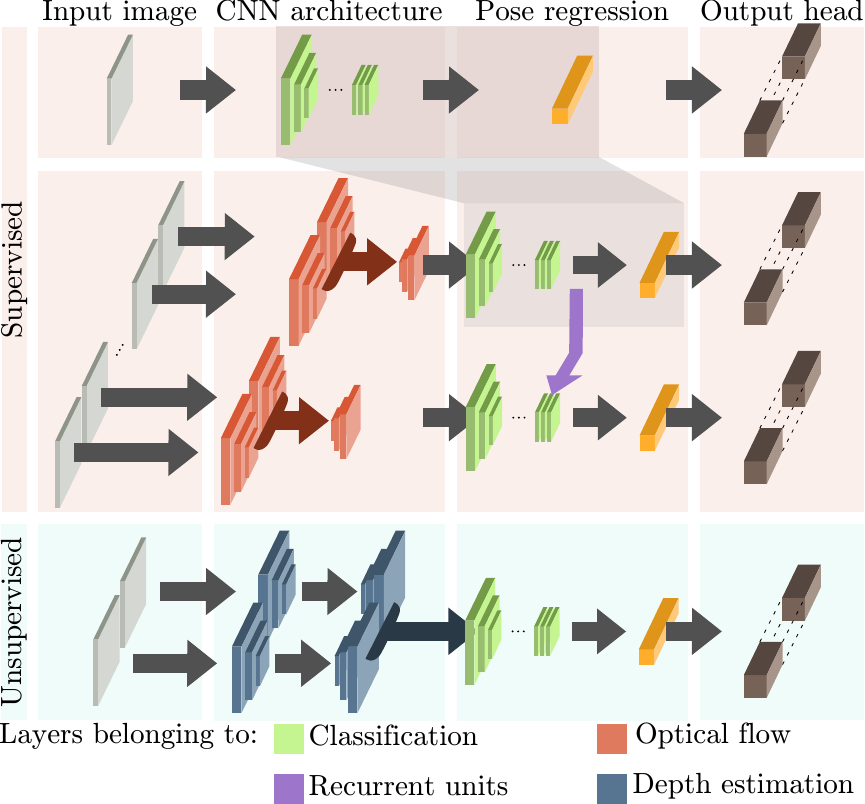}
  \caption{General representation of the end-to-end front-end pipelines. Early supervised approaches used classification networks as a baseline. More recently, the output from optical flow networks has been used as input to the classification network that precedes the pose regressor. In addition, some implementations leverage recurrent units. Unsupervised methods mainly rely on monocular depth estimation networks. The output is a pose estimate that may have a separated or joint translation and rotation estimation. }
  \label{gif:end2end-frontend}
\end{figure}

This section refers to "end-to-end front-end" as the mechanism that involves both appearance-based and geometry-based localization tasks in an end-to-end pipeline: given a query image (or sequence of images), it returns the predicted camera pose. It can also be referred to as end-to-end localization or end-to-end visual odometry.
They can be further divided into supervised and unsupervised according to the labels used during training, which also conditions the general formulation of the pipeline as shown in Fig. \ref{gif:end2end-frontend}. Following that, this section introduces supervised and unsupervised methods for end-to-end front-end implementations.

\subsubsection{Supervised methods}
\label{sec:frontend:vo:supervised}
The first end-to-end approach for camera localization involving loop detection and 6-DOF pose estimation was proposed in PoseNet \cite{kendall2015posenet}. It leverages the pre-trained classification network GoogleNet \cite{szegedy2015googlenet} as the baseline for pose regression, yielding an appearance-based method. More recent implementations of learning-based supervised methods for visual  odometry encode the appearance as optical flow fields, which can be represented in the image space \cite{wang2017deepvo,wang2020tartanvo}, or in a lower-dimensional latent space \cite{costante2018lsvo}. 
Moreover, in visual odometry, there is a strong temporal correlation between the frames' appearance. DeepVO \cite{wang2017deepvo} is constituted by a pre-trained FlowNet \cite{dosovitskiy2015flownet} \ac{CNN} preceded by two \ac{LSTM} units. The \ac{CNN} module models the appearance, and the recurrent unit models the dynamic relationships between sequences of images. 3DC-VO models the temporal relationships as a 3D CNN \cite{dl:vo:3dcvo}.
DeepVO does not require to be fed additional parameters for the camera model or the absolute scale, as the network learns them. However, this implies that the network can only be deployed under the data conditions it has been trained for. 
Conversely, TartanVO \cite{wang2020tartanvo} inputs the intrinsic camera parameters into a layer between the optical flow and pose retrieval modules to achieve generalization across different cameras. It implements PWC-Net \cite{sun2018pwcnet} for optical flow and a modified ResNet50 \cite{dl:vo:2015resnet} for pose retrieval, with two output heads for rotation and translation. A ResNet is also used to regress the pose in AtLoc \cite{dl:vo:wang2020atloc}, which additionally incorporates contextual information as a self-attention mechanism. The attention mechanism in \cite{dl:vo:xue2019guidedatt} operates in both the spatial and temporal domain, allowing it to detect stable features and to identify motion patterns. 
In addition to incorporating optical flow layers for matching, other algorithms leverage depth estimation modules. These modules can be either learned components \cite{czarnowski2020deepfactors, zhou2018deeptam} or geometric-based components \mbox{\cite{dl:vo:droidslam}}. This integrated approach enables the concurrent optimization of pose and depth estimation outputs, thereby enhancing the overall accuracy of the results.

The inductive bias of classification networks used for pose regression such as ResNet is shift equivariance \cite{dl:vo:cotogni2022offset}. This inductive bias makes pose regression techniques more closely related to image retrieval than to 3D geometry approaches, and it does not guarantee to generalize beyond the training data \cite{dl:vo:19cnnlimitations}. This realization suggests the need for leveraging networks that carry more geometric information.
The use of translation and rotation-equivariant features can induce the motion representation into the feature space, thus alleviating the amount of data needed and allowing for more lightweight models \cite{dl:vo:equivfeaturesforposeregression,dl:vo:zhang2020rotationequivariance}. Furthermore, the work in \cite{dl:vo:brynte2022rigidity} proposes an inductive bias with pitch-yaw equivariance which can handle perspective effects. 

Supervised pose regression networks usually rely on a Euclidean loss for rotation and translation \cite{kendall2015posenet, wang2017deepvo,dl:vo:wang2020atloc,dl:vo:xue2019guidedatt}. TartanVO trains the optical flow jointly with the pose regressor. To keep the loss consistent across different data sources, the pose loss proposes an up-to-scale function, which consists in normalizing the vector of the estimated translation.
The main challenge to address when implementing loss functions is the rotation representation.
PoseNet represents rotations with quaternions: any arbitrary quaternion can be easily mapped to a valid rotation via normalization. DeepVO chooses Euler angles to avoid the optimization restriction that the unit constraint for the quaternion implies. Both quaternions and Euler angles define a discontinuous representation for rotations. Neural networks tend to produce significant errors around discontinuities. Therefore, it is convenient to use smoother functions, or a continuous representation for rotations \cite{zhou2019continuity, dl:vo:isr}. The Lie algebra representation for the pose leverages an unconstrained optimization problem, which has proven to provide faster convergence when applied to PoseNet \cite{dl:frontend:lieposenet} and DeepVO \cite{dl:vo:deepvo-graph,dl:vo:isr}. 

\subsubsection{Unsupervised methods}

Most unsupervised methods rely on stereo-image pairs for training, and share a similar architecture that retrieves the scene depth and the pose estimate.

In UnDeepVO \cite{li2018undeepvo}, the pose estimator is a \ac{CNN} architecture that retrieves pose with orientation in Euler angles. The pose regressor consists of a VGG-based \cite{dl:loopclosure:vgg16} \ac{CNN} architecture. The depth estimator is an encoder-decoder network which directly recovers the depth map, rather than the inverse depth map, since it provides better convergence. 
D3VO \cite{yang2020d3vo} builds on top of the DepthNet network from MonoDepth2 \cite{godard2019monodepth2} and PoseNet for depth and pose estimation. SIMVODIS++\cite{kimICRA22simvodis++} can train depth and pose in a self-supervised manner using the attention module CBAM \cite{woo2018cbam}, applied to a sequence of three images.
Instead of stereo image pairs, SfMLearner \cite{zhou2017unsupervisedmono} can train a network that jointly learns depth and ego-motion from monocular video, both designed under a CNN framework. 

Unsupervised pose regression networks primarily rely on shift-equivariant CNN architectures \cite{dl:vo:cotogni2022offset}. The same principle to supervised approaches applies: due to the three-dimensional nature of the problem, these architectures could benefit from networks leveraging higher-dimensional inductive biases. For example, the aforementioned translation and rotation equivariant networks and the pitch-yaw equivariant networks \cite{dl:vo:equivfeaturesforposeregression,dl:vo:brynte2022rigidity}.

Without the ground-truth pose for the image sequences, the loss functions in unsupervised methods are based on variants of the photometric error.
UnDeepVO separates the loss into spatial and temporal components. The spatial loss operates with the stereo pair and minimizes the photometric error, checks the disparity map (inverse depth), and the pose consistency between the stereo pair images. The temporal loss operates with monocular images and minimises the photometric and 3D geometric registration errors.
D3VO depth and pose estimation are jointly trained using a photometric constancy loss function. Moreover, the loss function integrates uncertainty terms trained to model illumination changes or noise sources such as Lambertian surfaces.
In addition to the photometric loss, SIMVODIS++ leverages three other metrics: the structural similarity loss handles inconsistencies from non-Lambertian surfaces and illumination changes by calculating the similarity between image patches. The depth smoothness loss removes random sharp lines, and the field-of-view loss uses the \ac{MSE} for learning the camera parameters.

One limitation of the photometric loss is the drift caused by scale consistency for depth across images. Trianflow aligns the scale of depth and poses with a two-view triangulation \cite{zhao2020trianflow}. SC-SfMLearner \cite{bian2019unsuperviseddepthconsistency} builds on top of SfMLearner, introducing a geometry consistency loss. It estimates the pose between consecutive frames using PoseNet, and the depth on each of them with DepthNet. Afterwards, it warps one depth frame onto another using the estimated transform, leveraging the geometric consistency. This way, the scale remains consistent between consecutive frames.

\subsection{End-to-end back-end}
\label{sec:deeplearning:backend}
End-to-end back-end approaches exploit the sparsity of the \ac{SLAM} graph by leveraging \ac{GNN}s.
The work in \cite{tanaka2021learningbackend} builds a \ac{GNN} in which the vertices comprise the camera poses (keyframes) and landmarks, and the edges connect them according to visibility. Similar to \ac{BA}'s implementation in ORB-SLAM, the sum of the reprojection errors with the Huber norm is used as a loss function. While it does not outperform the standard \ac{BA} from the g2o library \cite{grisetti2011g2o} in terms of accuracy, it shows five times faster performance.

PoGO-Net \cite{li2021pogoNET} builds the graph upon a multiple rotation averaging approach with a quaternion parameterization. Therefore, the vertex represents the absolute camera orientations and the edges' pair-wise connections.
Furthermore, it addresses \ac{GNN}'s vulnerability to noisy graphs by adding parameterized denoising layers for outlier edge removal. The loss function has a local and a global consistency component. The global consistency component is used to train the \ac{GNN} and evaluates the precision of the pose graph. The local consistency component is used to evaluate the denoising layer, and it evaluates the prediction of the absolute camera orientations.
Instead of only making a binary vertex connection according to visibility, the edges are parameterized to provide a reliability metric for the connections.

A different approach is followed by RL-PGO\cite{kourtzanidis2022rlpgo}, which implements a reinforcement learning algorithm for a 2D case of pose graph optimization. It learns a policy for predicting optimal orientation retractions from pose graph observations. It outperforms Gauss-Newton and Levenberg-Marquardt methods, although it does not scale well with the size of the image and therefore is not suitable for online implementations.

The recent  neural radiance field (NeRF) architecture \cite{mildenhall2021nerf} has been effectively incorporated as the mapping module in iMaP \cite{sucar2021imap} and NeRF-SLAM \cite{RW:nerfslam}. These frameworks, when provided with an image and pose input, are capable of producing per-pixel depth and colour outputs.

\subsection{End-to-end SLAM}
\label{sec:deeplearning:endSLAM}

The main challenge to address in the end-to-end SLAM approaches is the development of a fully-differentiable pipeline. The high modularity of SLAM, with non-differentiable and complex subsystems, has hindered the development of end-to-end SLAM pipelines, which still remains an open problem. 

SLAM-net \cite{karkus2021SLAM-Net} encodes the particle filter implementation of FastSLAM \cite{montemerlo2002fastslam} to learn mapping, transition, and observation models. Each particle represents a trajectory hypothesis and an associated weight.
The network implements three models: a transition model, an observation model, and a mapping model. The outputs are the 2D pose estimate and a 2D grid map.
The transition model is a learnt CNN estimating relative motion between frames. It extends the particle trajectories with each new observation.
The mapping model is a learnt CNN that predicts the local map, which is added to a list of previously predicted local maps that configure a 2D grid of latent features.
The local maps and the trajectory hypotheses are fed into the observation model, which updates the trajectory weights according to the consistency between maps and trajectory.
The data transition from the mapping model to the observation model is made differentiable using spatial transformers \cite{jaderberg2015spatialtransformers} to convert the local map according to the relative pose in the particle trajectory.

In MapNet \cite{henriques2018mapnet}, camera localization and map are encoded in a 2.5D representation, in which information related to
the vertical dimension is implicitly encoded as a feature vector. The pose estimate is obtained from a standard convolutional operator as the absolute pose in the map. The registration of new observations to the map is obtained from its dual
operator, deconvolution. This way, the SLAM problem is formulated as a simple and differentiable pipeline. 

The non-differentiability of raycasting techniques required in the 3D to 2D mapping and some optimization algorithms used in SLAM such as Levenberg-Marquardt leads to simplified formulations for end-to-end SLAM where the map is only represented in the latent space. The work in gradSLAM \cite{jatavallabhula2020gradslam} presents a framework for developing fully differentiable SLAM systems. Within this framework, differentiable trust-region optimizers, map fusion schemes, and ray backprojection from the inverse depth map to the 2D image have been implemented. 

Although the development of end-to-end SLAM is still in its early stages, these recent advances present promising lines of development for future architectures. 

\begin{table*}[!htbp]
\caption{Comparison of the state-of-the-art  datasets for SLAM.}
\centering
\scriptsize
\label{table:comparisonsoadataset}
\begin{tabular}{|c|cc cc cc cc c| }
\hline
Dataset                                   & Type       & Environment   & IMU      & Segmentation & Camera    & Depth   & Perceptual aliasing  & Dynamic &  Varying lighting\\    
\hline \hline
 
 EuRoC \cite{burri2016euroc}       & real      & indoor        & \cmark    & \xmark       & stereo    & \xmark  & \xmark               & \xmark  & \cmark\\ 
 TUM-RGBD \cite{sturm2012tumrgbd} & real      & indoor        & \cmark    & \xmark       & stereo    & \cmark  & \cmark                  & \cmark & \xmark \\ 
 RIO10 \cite{wald2019rio10} & real      & indoor        & \xmark    & \cmark       & monocular    & \cmark  & \xmark                  & \xmark & \cmark\\  
  OpenLORIS \cite{openloris} & real      & indoor        & \cmark    & \xmark       & monocular    & \cmark  & \xmark                  & \xmark & \xmark \\  
  BONN-RGBD \cite{palazzolo2019bonndynamic} & real      & indoor        & \xmark    & \xmark       & monocular    & \cmark  & \xmark                  & \cmark & \xmark \\ 
 ETH-MS \cite{eth_ms_visloc_2021}  & real      & in/outdoor    & \xmark    & \xmark       & rig       & \xmark  & \xmark               & \xmark  & \cmark \\ 
  KITTI \cite{dataset:kitti}                & real      & city          & \cmark    & \cmark       & stereo    & \cmark  & \xmark               & \cmark  & \cmark \\  
 RobotCar Seasons \cite{maddern2017Robotcar} & real      & city        & \cmark    & \xmark       & monocular    & \xmark  & \cmark                  & \xmark & \cmark \\ 
  UMCD \cite{UMCD} & real      & city        & \cmark    & \xmark       & monocular    & \xmark  & \cmark                  & \xmark & \cmark \\ 
 Aqualoc \cite{dataset:aqualocdb}                  & real      & underwater    & \cmark    & \xmark       & monocular & \xmark  & \cmark               & \cmark  & \cmark\\
 AURORA \cite{bernardi2022aurora}  & real      & underwater    & \cmark    & \xmark       & monocular & \xmark  & \cmark               & \cmark & \cmark\\
TartanAIR\cite{tartanair2020iros}  & synthetic & miscellaneous & \cmark    & \cmark       & monocular       & \cmark  & \xmark                 & \cmark & \cmark \\
MIMIR-UW \cite{dataset:mimir} & synthetic & underwater    & \cmark    & \cmark       &rig        & \cmark  & \cmark               & \cmark & \cmark\\    
\hline

\end{tabular}
\end{table*}

\section{Environment-specific challenges}
\label{sec:environment}
One of SLAM’s most severe challenges is developing robust architectures
being applied in various working conditions without compromising accuracy under various sources of degradation. 

Indirect methods are sensitive to geometrically-degraded environments by textureless or repetitive patterns, that is, perceptual aliasing. 
The lack of features and the wrong matching between them lead to wrong tracking or tracking failure. Direct methods are more sensitive than indirect methods to high parallax, due to the higher nonconvexity of the optimization. In both cases, visually degraded environments such as those with low light, illumination changes and Lambertian surfaces are known sources of tracking failure. 

Both geometrically and visually degraded images by dusty or underwater environments, or from dynamic elements in the scene, are another source of spurious estimates. For that case, learning-based pipelines that can encode higher-level representations of the semantic information are presented as an alternative. On the other hand, a limitation exclusive to learning-based methods is the variety of imaging conditions and motion patterns, which affects their generalizability. The remainder of this section presents an experimental evaluation of the aforementioned challenges, as detailed below.

\subsection{Experimental evaluation}
This section carries out an experimental evaluation of the SLAM algorithms surveyed throughout the paper \footnote{The code for the experimental evaluation is available at \url{https://github.com/olayasturias/monocular_visual_slam_survey}}. Moreover, the test cases are selected according to the challenges mentioned above.

\subsubsection{Test datasets}
\begin{figure*}
\centering
\resizebox{\textwidth}{!}{\begin{tabular}{p{.2cm}c@{\hspace{.7mm}}c@{\hspace{3mm}\color{ white}\vrule}!{\color{ white}\vrule}c@{\hspace{.7mm}}c@{\hspace{3mm}\color{ white}\vrule}!{\color{ white}\vrule}c@{\hspace{.7mm}}c@{\hspace{3mm}\color{ white}\vrule}!{\color{ white}\vrule}c@{\hspace{.7mm}}c@{\hspace{3mm}}}

 &\multicolumn{2}{c@{\hspace{3mm}}}{00}
 &\multicolumn{2}{c@{\hspace{3mm}}}{01}
 &\multicolumn{2}{c@{\hspace{3mm}}}{02}
 &\multicolumn{2}{c@{\hspace{3mm}}}{03}\\

\rotatebox[origin=c]{90}{KITTI \cite{dataset:kitti}} 
&\multicolumn{2}{c@{\hspace{3mm}}}{\adjustbox{valign=c,vspace=.2pt }{\includegraphics[width=.4\linewidth ]{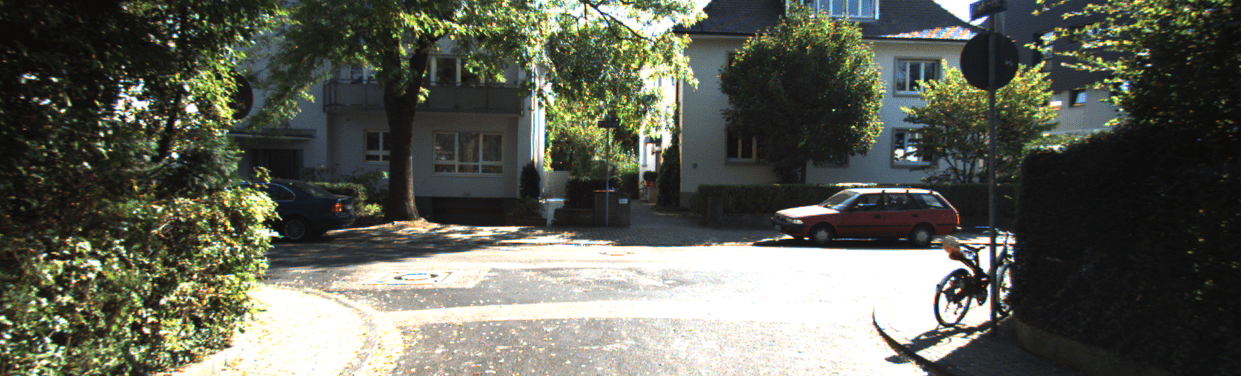}}}
&\multicolumn{2}{c@{\hspace{3mm}}}{\adjustbox{valign=c,vspace=.2pt }{\includegraphics[width=.4\linewidth ]{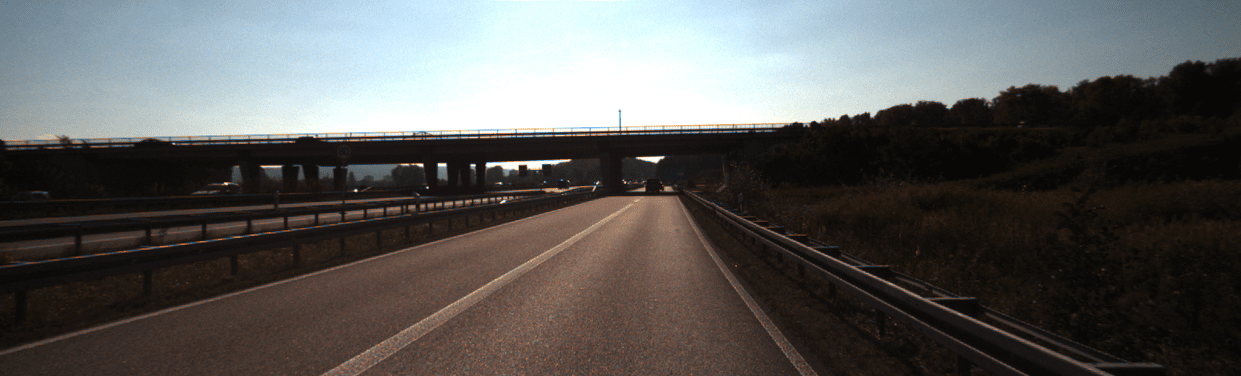}} }
&\multicolumn{2}{c@{\hspace{3mm}}}{\adjustbox{valign=c,vspace=.2pt }{\includegraphics[width=.4\linewidth ]{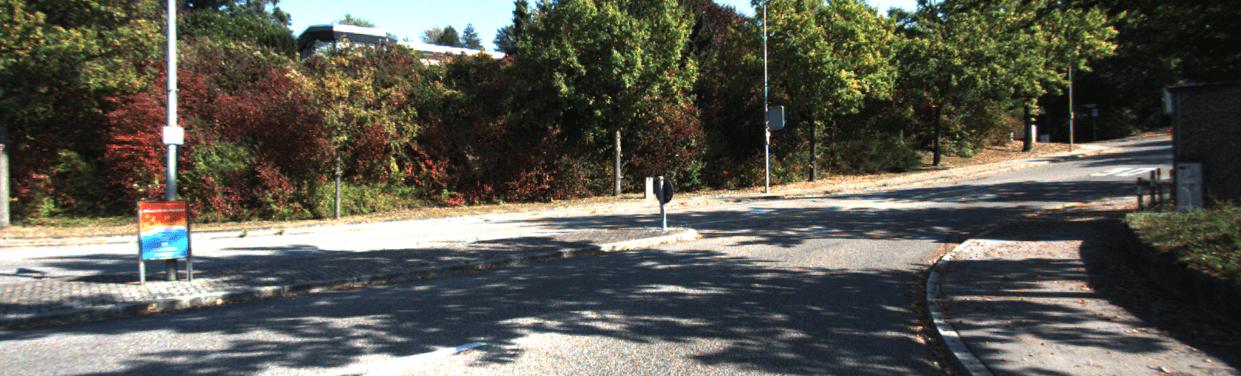}} }
&\multicolumn{2}{c@{\hspace{3mm}}}{\adjustbox{valign=c,vspace=.2pt }{\includegraphics[width=.4\linewidth ]{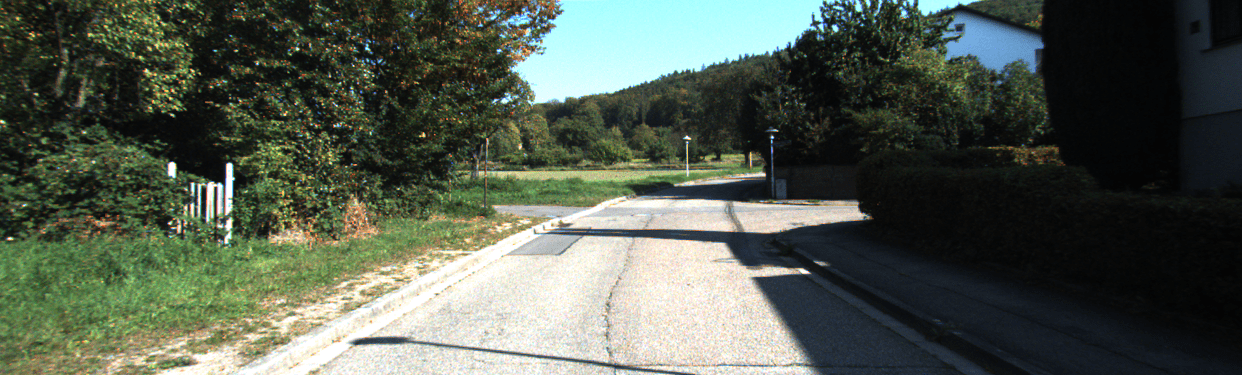}} }\\

 &\multicolumn{2}{c@{\hspace{3mm}}}{MH\_01\_easy}
 &\multicolumn{2}{c@{\hspace{3mm}}}{MH\_04\_difficult}
 &\multicolumn{2}{c@{\hspace{3mm}}}{V1\_02\_medium} &\multicolumn{2}{c@{\hspace{3mm}}}{V1\_03\_difficult} \\ 
 
\rotatebox[origin=c]{90}{EuRoC \cite{burri2016euroc}} 
& \adjustbox{valign=m,vspace=.2pt}{\includegraphics[width=.2\linewidth]{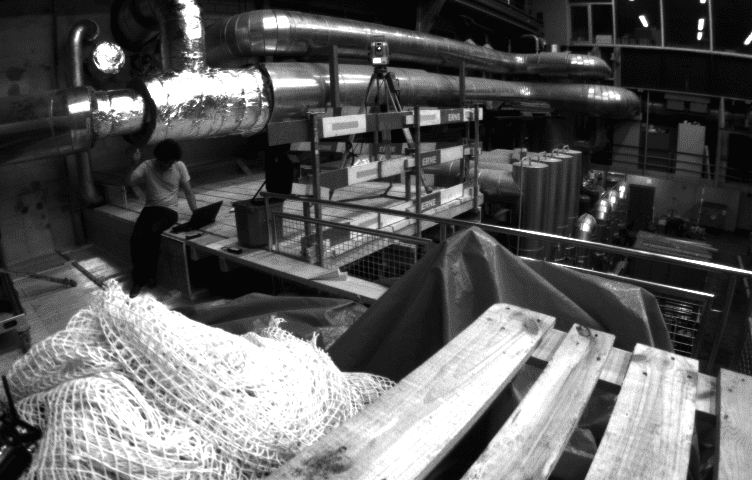}} 
& \adjustbox{valign=m,vspace=.2pt}{\includegraphics[width=.2\linewidth]{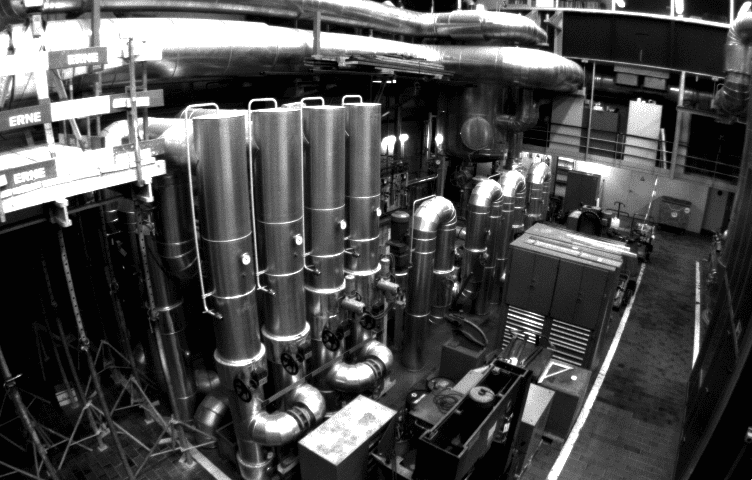}}

& \adjustbox{valign=m,vspace=.2pt}{\includegraphics[width=.2\linewidth]{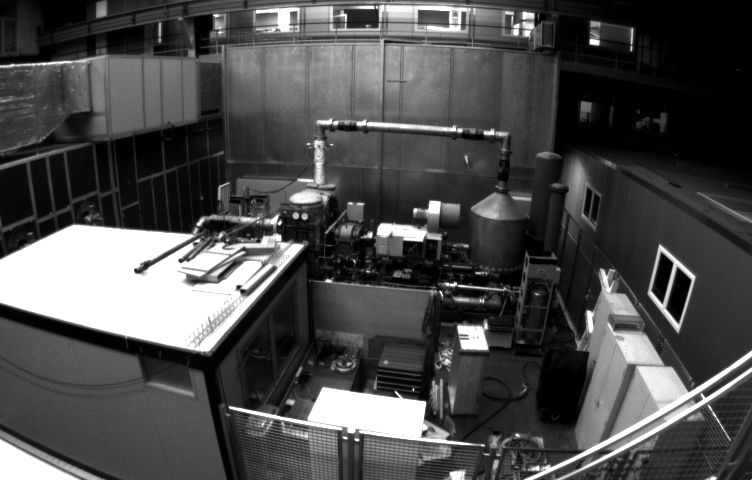}}
& \adjustbox{valign=m,vspace=.2pt}{\includegraphics[width=.2\linewidth]{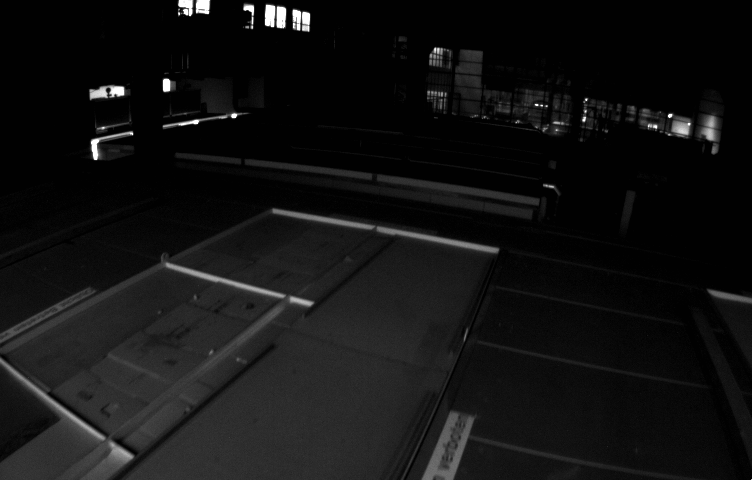}}

& \adjustbox{valign=m,vspace=.2pt}{\includegraphics[width=.2\linewidth]{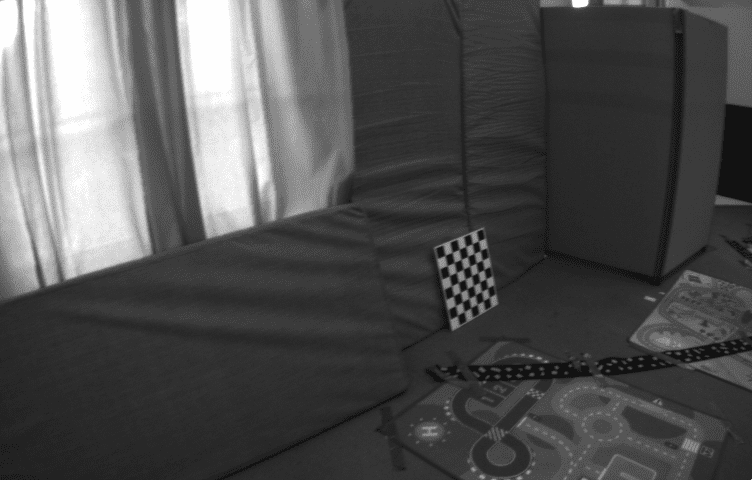}}
& \adjustbox{valign=m,vspace=.2pt}{\includegraphics[width=.2\linewidth]{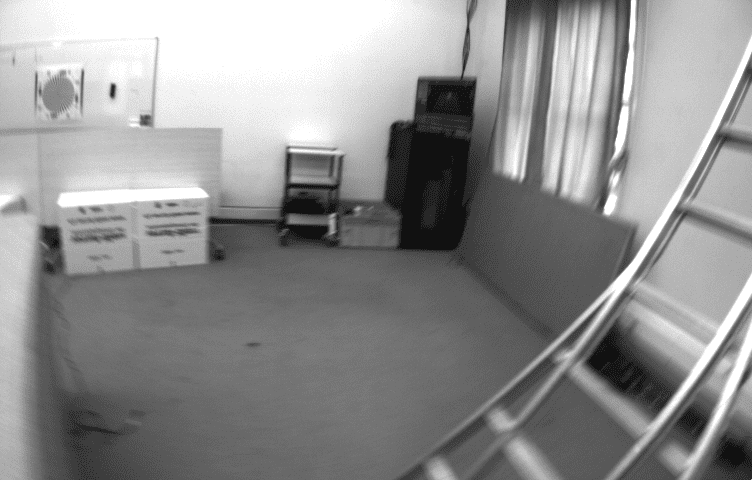}}

& \adjustbox{valign=m,vspace=.2pt}{\includegraphics[width=.2\linewidth]{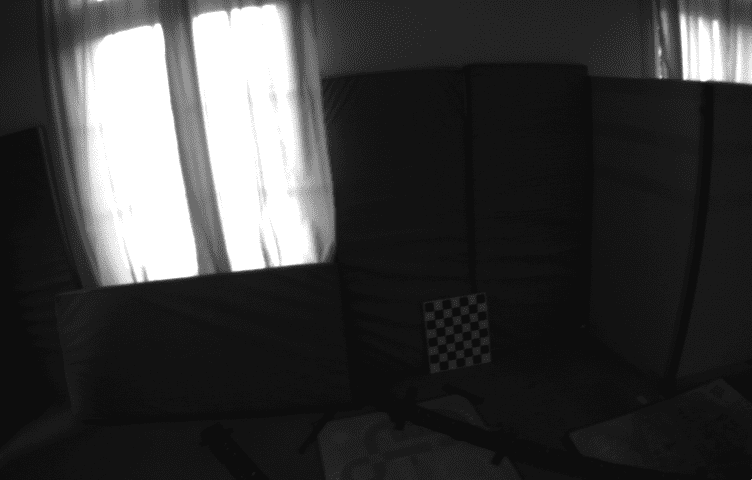}}
& \adjustbox{valign=m,vspace=.2pt}{\includegraphics[width=.2\linewidth]{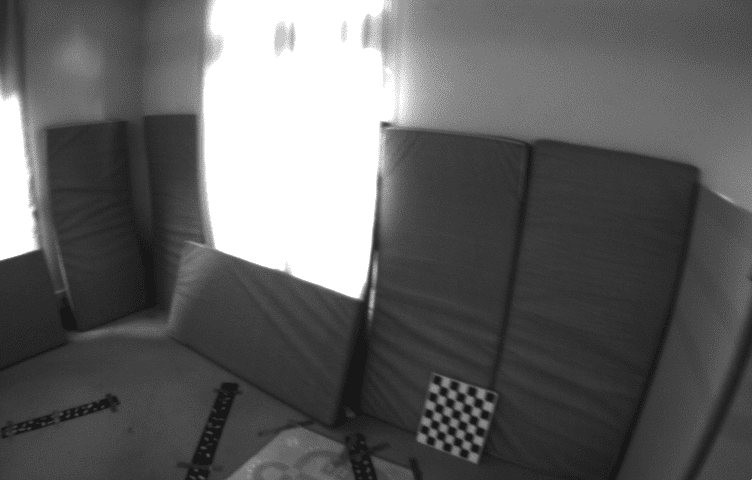}}\\

 & \multicolumn{2}{c@{\hspace{3mm}}}{fr1/360}
 & \multicolumn{2}{c@{\hspace{3mm}}}{fr1/rpy}
 &  \multicolumn{2}{c@{\hspace{3mm}}}{fr3/nostructure\_notexture\_far}
 &  \multicolumn{2}{c@{\hspace{3mm}}}{fr3/nostructure\_texture near\_loop} \\ 
 
\rotatebox[origin=c]{90}{TUM-RGBD \cite{sturm2012tumrgbd}} 
& \adjustbox{valign=m,vspace=.2pt}{\includegraphics[width=.2\linewidth]{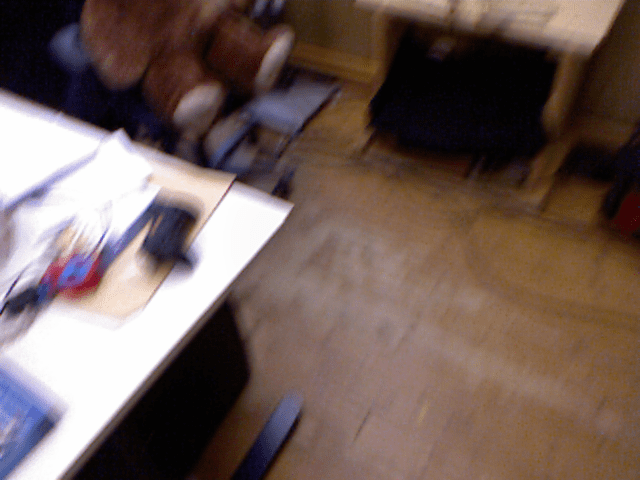}} 
& \adjustbox{valign=m,vspace=.2pt}{\includegraphics[width=.2\linewidth]{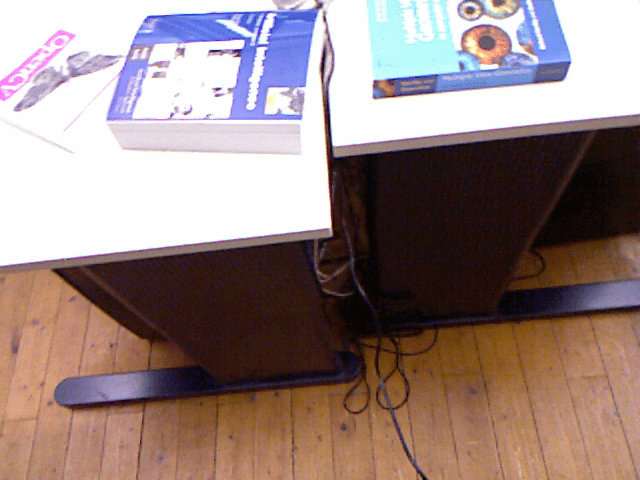}}

& \adjustbox{valign=m,vspace=.2pt}{\includegraphics[width=.2\linewidth]{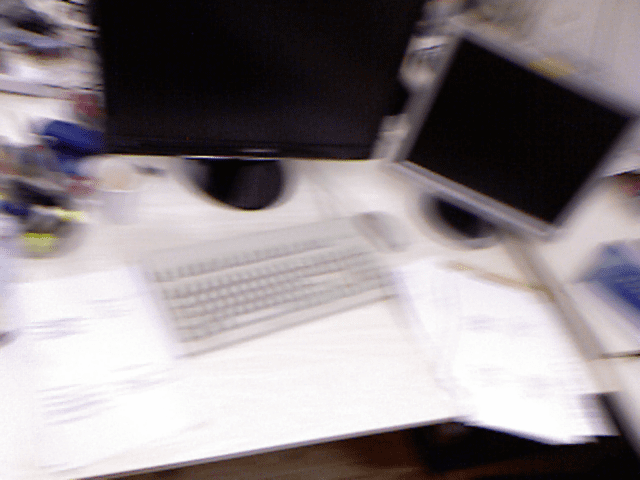}}
& \adjustbox{valign=m,vspace=.2pt}{\includegraphics[width=.2\linewidth]{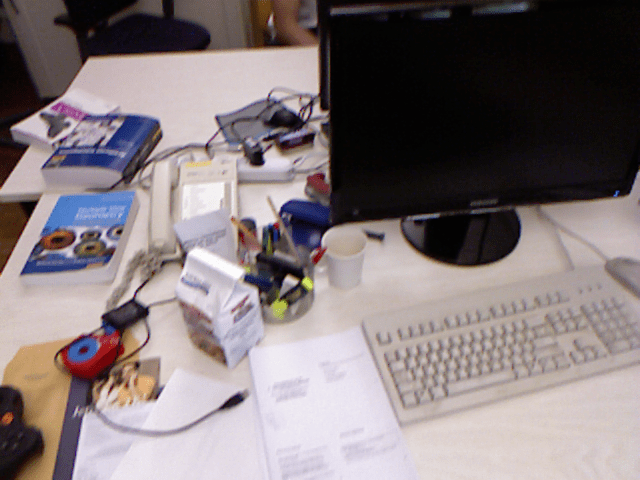}}

& \adjustbox{valign=m,vspace=.2pt}{\includegraphics[width=.2\linewidth]{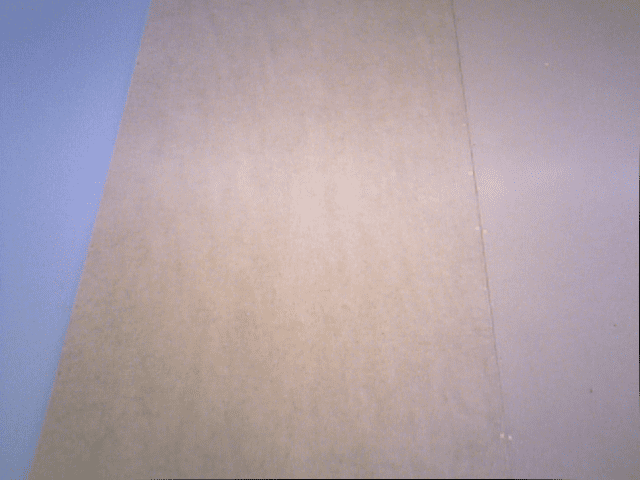}}
& \adjustbox{valign=m,vspace=.2pt}{\includegraphics[width=.2\linewidth]{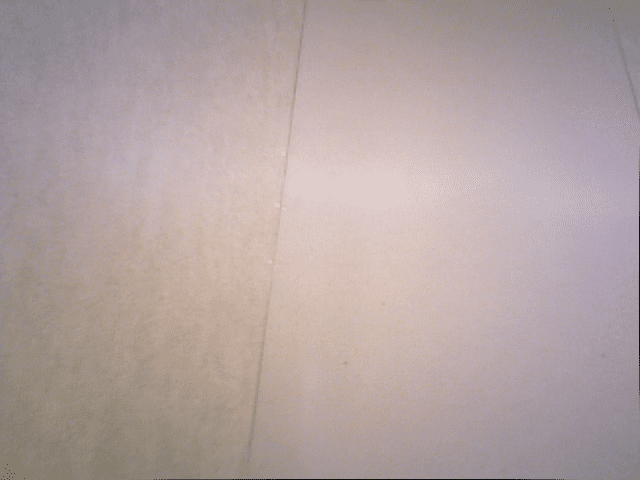}}

& \adjustbox{valign=m,vspace=.2pt}{\includegraphics[width=.2\linewidth]{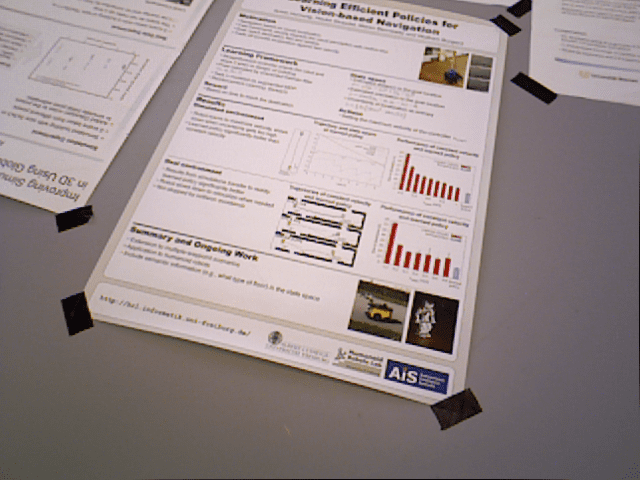}}
& \adjustbox{valign=m,vspace=.2pt}{\includegraphics[width=.2\linewidth]{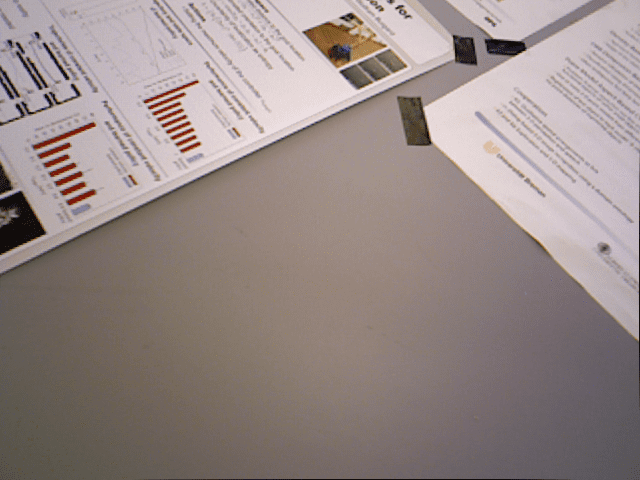}}\\

 & \multicolumn{2}{c@{\hspace{3mm}}}{Archaeological sequence 1} & \multicolumn{2}{c@{\hspace{3mm}}}{Archaeological sequence 2} &  \multicolumn{2}{c@{\hspace{3mm}}}{Archaeological sequence 3} &  \multicolumn{2}{c@{\hspace{3mm}}}{Archaeological sequence 4} \\ 
 
\rotatebox[origin=c]{90}{Aqualoc \cite{dataset:aqualocdb}} 
& \adjustbox{valign=m,vspace=.2pt}{\includegraphics[width=.2\linewidth]{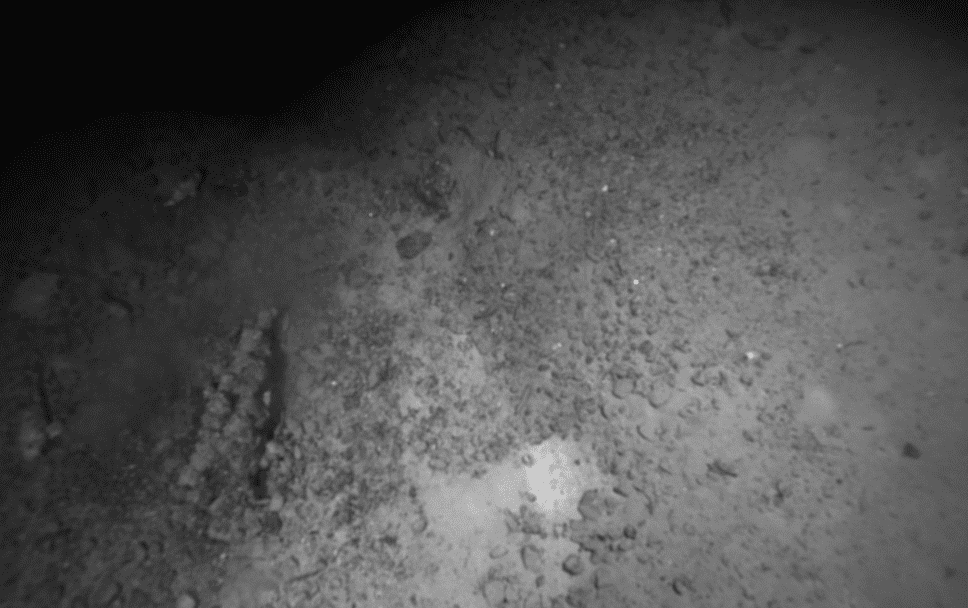}} 
& \adjustbox{valign=m,vspace=.2pt}{\includegraphics[width=.2\linewidth]{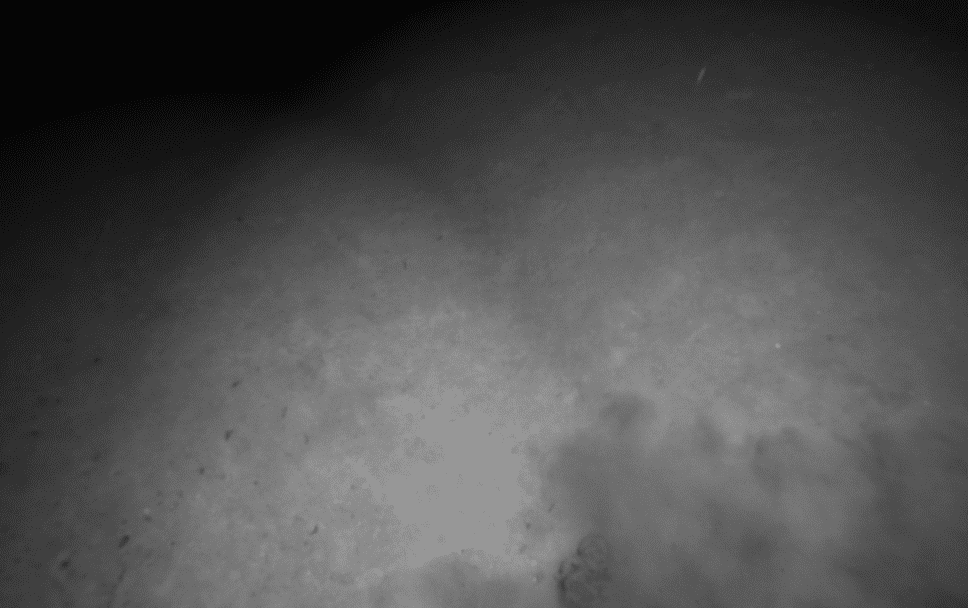}}

& \adjustbox{valign=m,vspace=.2pt}{\includegraphics[width=.2\linewidth]{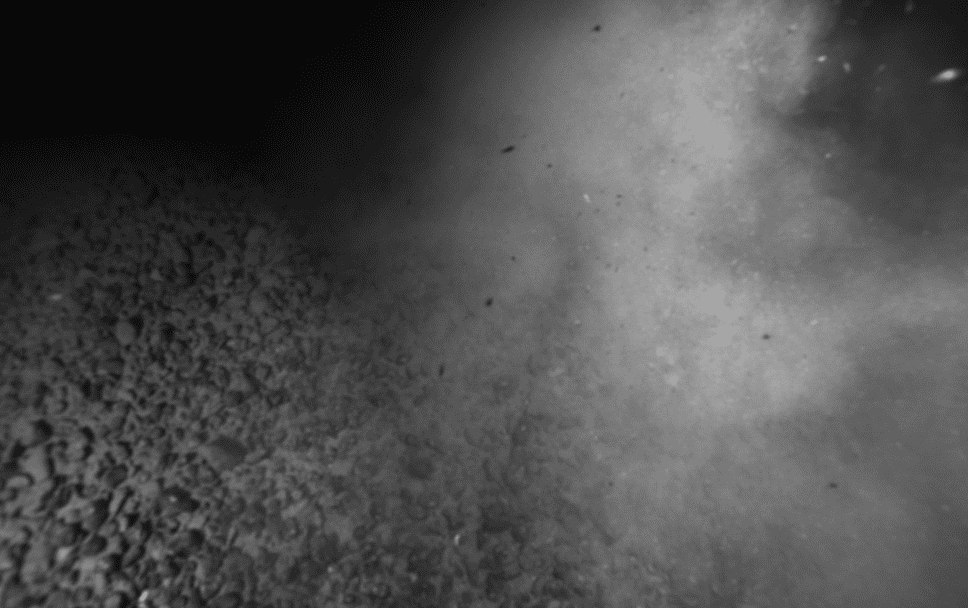}}
& \adjustbox{valign=m,vspace=.2pt}{\includegraphics[width=.2\linewidth]{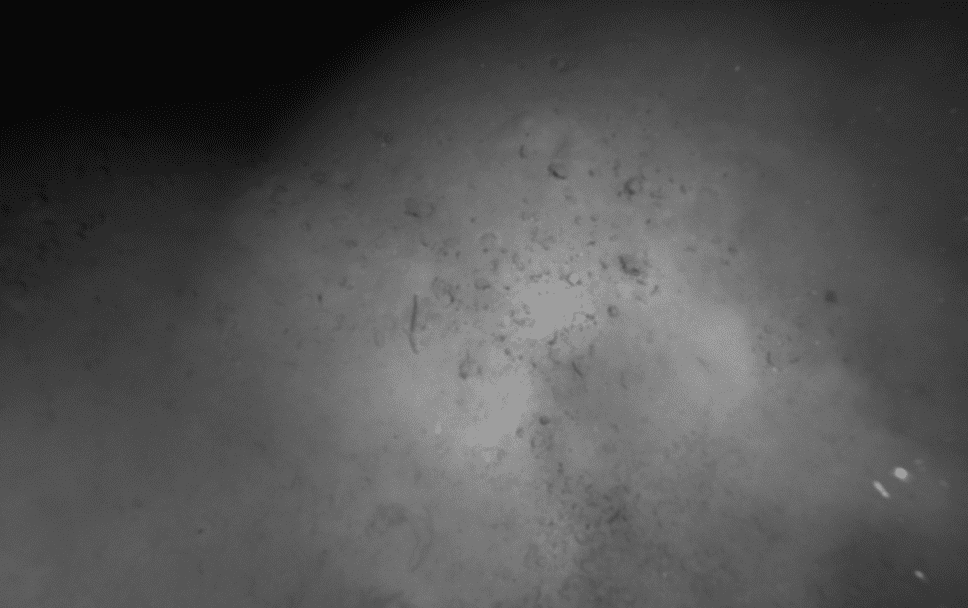}}

& \adjustbox{valign=m,vspace=.2pt}{\includegraphics[width=.2\linewidth]{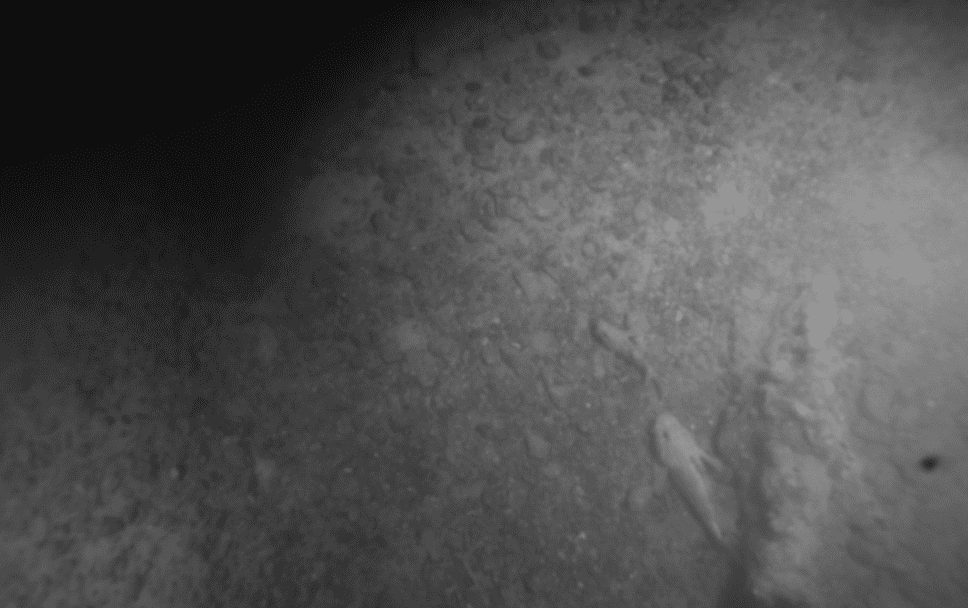}}
& \adjustbox{valign=m,vspace=.2pt}{\includegraphics[width=.2\linewidth]{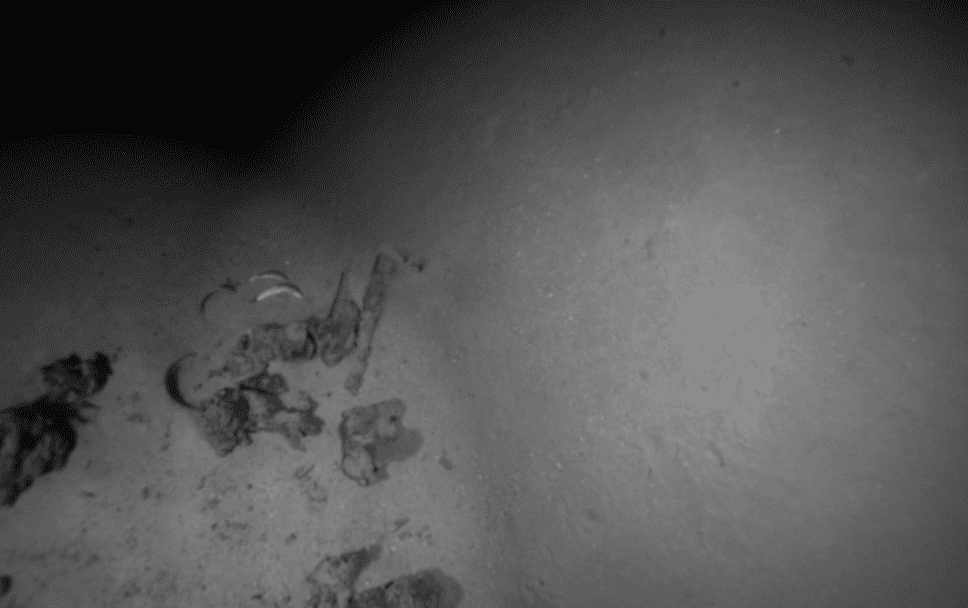}}

& \adjustbox{valign=m,vspace=.2pt}{\includegraphics[width=.2\linewidth]{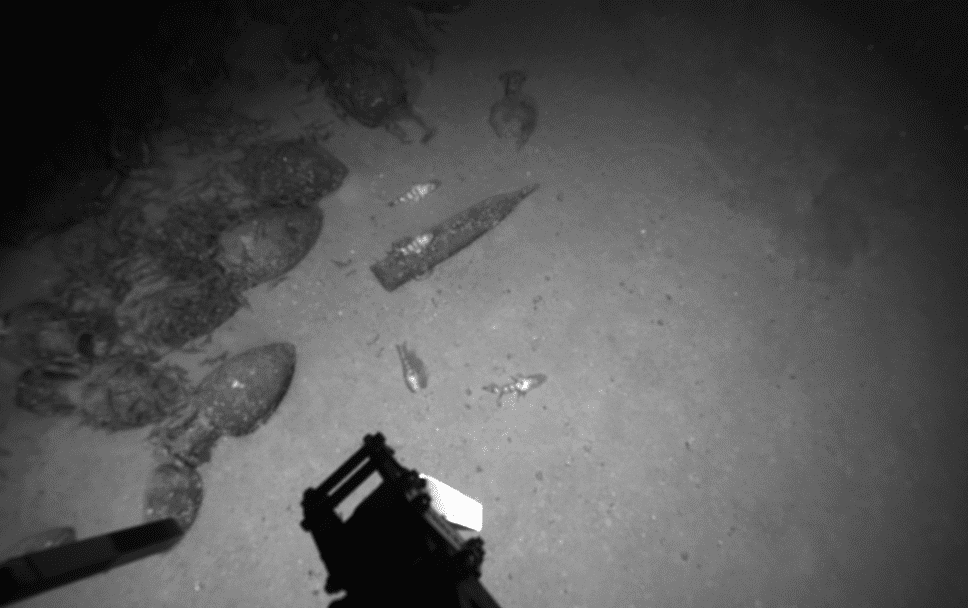}}
& \adjustbox{valign=m,vspace=.2pt}{\includegraphics[width=.2\linewidth]{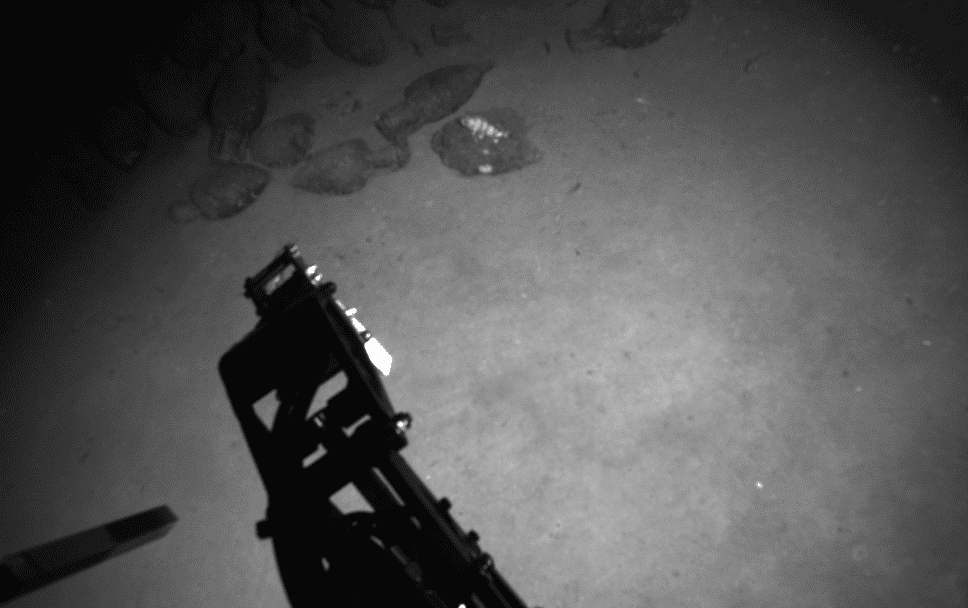}}\\

 & \multicolumn{2}{c@{\hspace{3mm}}}{SeaFloor} & \multicolumn{2}{c@{\hspace{3mm}}}{SeaFloor\_Algae} &  \multicolumn{2}{c@{\hspace{3mm}}}{OceanFloor} &  \multicolumn{2}{c@{\hspace{3mm}}}{SandPipe} \\ 
 
\rotatebox[origin=c]{90}{MIMIR-UW \cite{dataset:mimir}} 
& \adjustbox{valign=m,vspace=.2pt}{\includegraphics[width=.2\linewidth]{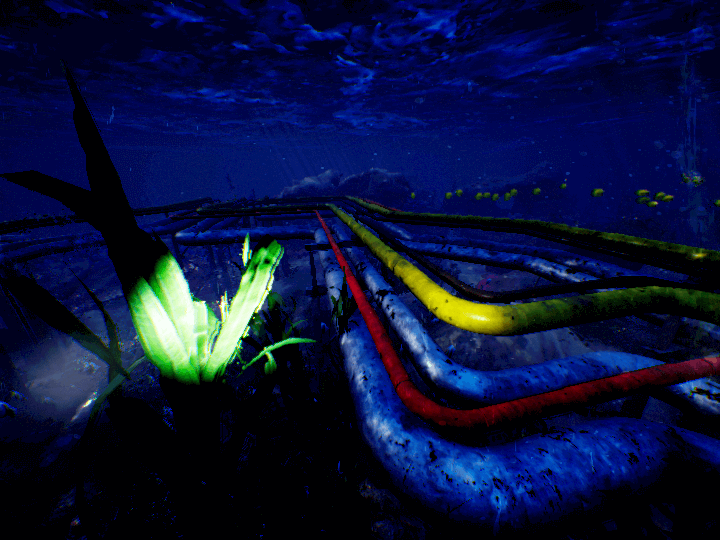}} 
& \adjustbox{valign=m,vspace=.2pt}{\includegraphics[width=.2\linewidth]{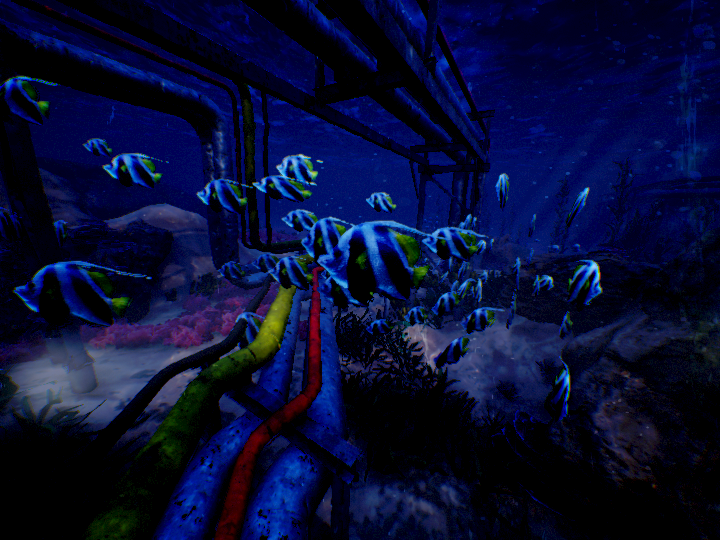}}

& \adjustbox{valign=m,vspace=.2pt}{\includegraphics[width=.2\linewidth]{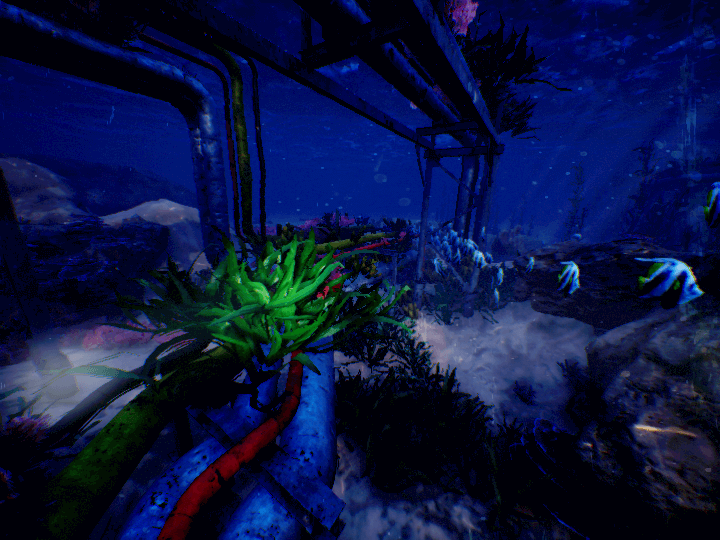}}
& \adjustbox{valign=m,vspace=.2pt}{\includegraphics[width=.2\linewidth]{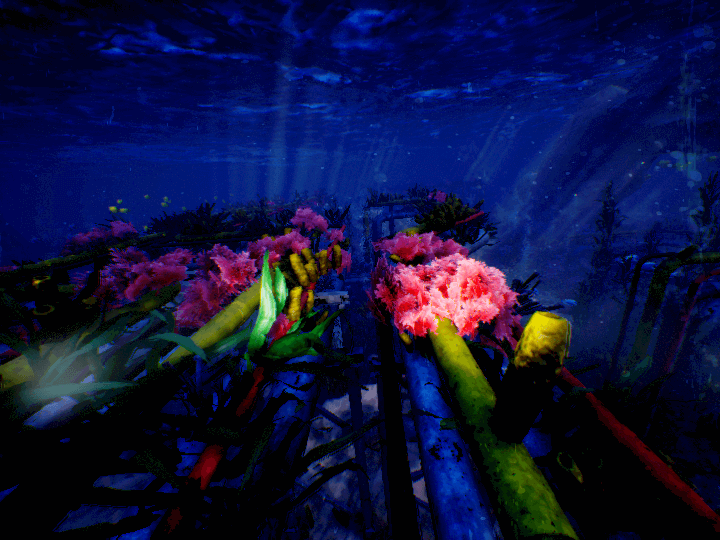}}

& \adjustbox{valign=m,vspace=.2pt}{\includegraphics[width=.2\linewidth]{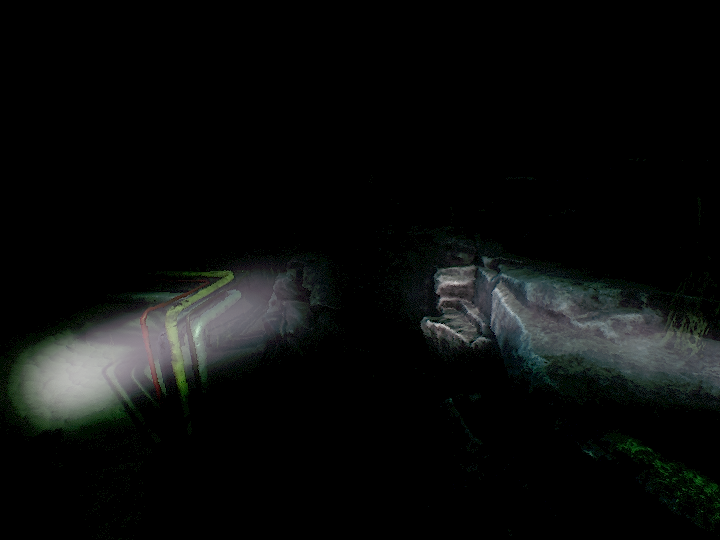}}
& \adjustbox{valign=m,vspace=.2pt}{\includegraphics[width=.2\linewidth]{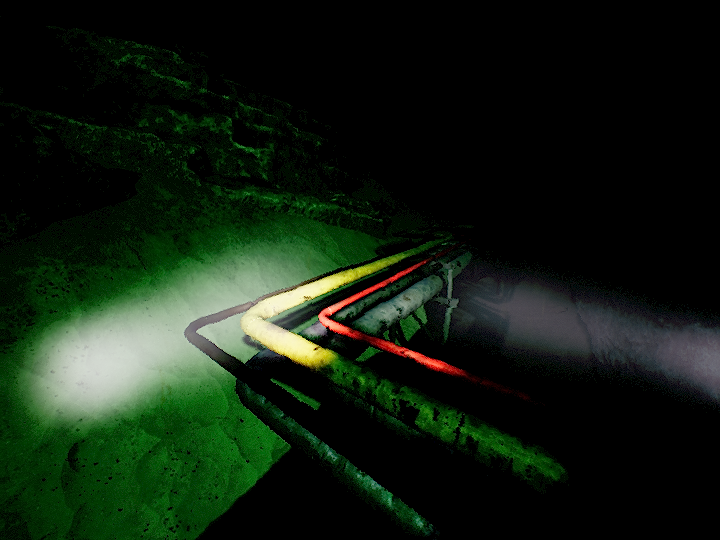}}

& \adjustbox{valign=m,vspace=.2pt}{\includegraphics[width=.2\linewidth]{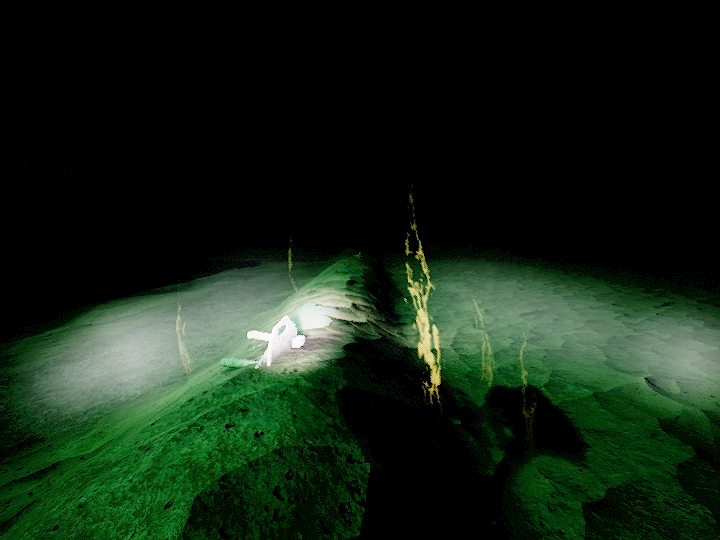}}
& \adjustbox{valign=m,vspace=.2pt}{\includegraphics[width=.2\linewidth]{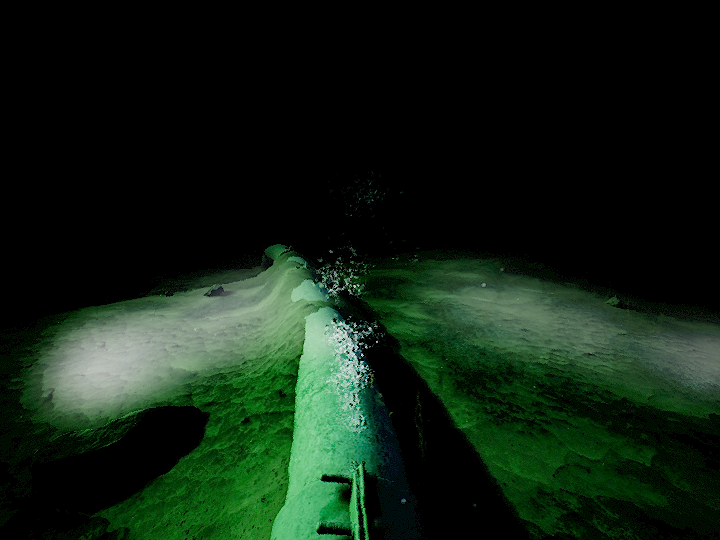}}\\

\end{tabular}}
\caption{Images from the datasets chosen for experimental evaluation. The roads where KITTI is recorded have varying degrees of structural elements. EuRoC assigns difficulty levels based on the presence of lighting changes and motion blur. TUM-RGBD displays motion blur and image sequences with no texture. Aqualoc and MIMIR-UW measurements are taken underwater, with uneven illumination from artificial lightning, dynamic elements, and floating particles. }
\label{fig:sampleimgs}
\end{figure*}

Table \ref{table:comparisonsoadataset} lists some prevailing SLAM databases and the challenges for SLAM they present. A sample of these datasets has been selected for the experimental evaluation, such that the SLAM algorithms are tested under various environment-specific challenges. This sample consists of the following datasets: KITTI \cite{dataset:kitti}, EuRoC \cite{burri2016euroc}, TUM-RGBD \cite{sturm2012tumrgbd}, Aqualoc \cite{dataset:aqualocdb}, and MIMIR-UW \cite{dataset:mimir}. Their imaging conditions are exemplified in Fig. \ref{fig:sampleimgs}.

KITTI is recorded from a car, which severely limits the motion patterns of the sequences. It contains sequences with dynamic elements (e.g., moving cars or pedestrians), structured scenes from cities or motorways, and unstructured scenes from rural areas. EuRoC \cite{burri2016euroc} is recorded with a drone, containing sequences recorded under different difficulty levels. The difficult sequences for "machine hall" include motions at high speeds, with the subsequent motion blur, while the ones for "Vicon room" present significant illumination changes across frames. TUM-RGBD \cite{sturm2012tumrgbd} is a handheld-recorded dataset. It includes sequences recorded under pure rotations, with the subsequent view changes and motion blur, and a sequence recorded under the total absence of structure and texture. The two remaining datasets, Aqualoc and MIMIR-UW, are recorded underwater, with imaging conditions characteristic of these environments. Underwater environments commonly present repetitive patterns or untextured areas, which might, for example, come from the seabed's lack of structure, the presence of floating particles, and the backscattering effects.
Furthermore, they often present many dynamic elements, such as fishes, vegetation, and floating particles. The lack of natural light in underwater environments often requires robots to carry artificial light. This artificial lighting creates scattering effects from floating particles, uneven lighting across the scene, and illumination changes of the objects across frames. These challenges are present in Aqualoc and MIMIR-UW. 
MIMIR-UW is a simulated underwater dataset with dynamic elements, varied motion patterns, and lighting conditions, including artificial lights under very dark scenes. Aqualoc presents frames with uniform texture and dynamic elements from fishes, robot parts, and particle dust. It follows a very uniform horizontal motion pattern.

\subsubsection{Evaluation metrics}
The selected metrics are the standard accuracy metrics for trajectory estimates as implemented in \cite{grupp2017evo}. 
Let us assume an estimated trajectory $P_i \in SE_3$ and the ground truth $Q_i \in SE_3$ composing a sequence of time-synchronized spatial poses.

The \ac{APE} measures the Euclidean distance between the absolute poses at each timestamp $i$.
This metric quantifies the global consistency of the trajectory.  
Ground-truth and estimate coordinate frames are first aligned using the least-squares alignment implemented in \cite{grupp2017evo} to retrieve the transform $S \in Sim_3$ that best aligns $P_i$ with $Q_i$:
\begin{equation}
   APE_i= Q_i^{-1}SP_i 
\end{equation}

The \ac{RPE} quantifies the local consistency of the trajectory. It measures the Euclidean distance at time step $i$ between the consecutive poses over a fixed time interval $\Delta$, such  that:
\begin{equation}
    RPE_i = \left(Q_i^{-1}Q_{i+\Delta} \right)^{-1}\left(P_i^{-1}P_{i+\Delta} \right)
\end{equation}

For the overall trajectory, the \ac{APE} and \ac{RPE} are obtained as the \ac{RMSE} over all time intervals $\Delta$:

\begin{align}
   APE_{RMSE}&= \sqrt{\frac{1}{n} \sum^n_{i=1}||APE_i||^2} \\
    RPE_{RMSE} &= \sqrt{\frac{1}{m} \sum^m_{i=1}||RPE_i||^2}  ,
\end{align}
with $m = n-\Delta$. The translation and rotation components are computed separately. The rotation errors are obtained as the geodesic distance $||\log_{SO3}\left((E_i)\right)||$, with $E_i$ the representing either the $APE_i$ or the $RPE_i$, and $\log_{SO3}(\cdot)$ the logarithmic mapping of $SO3$ \cite{huynh2009metrics}.

\subsubsection{Evaluation results}
The open-source algorithms ORB-SLAM3, DSO, DF-VO,  and TrianFlow  are tested in the state-of-art mentioned above datasets. ORB-SLAM2 and DSO are indirect and direct approaches to geometry-based SLAM. DF-VO represents a hybrid architecture, with a learning-based pipeline trained with KITTI for optical flow and depth estimation, and a geometry-based PnP algorithm. TrianFlow is an end-to-end learning-based pipeline trained under KITTI and TUM\_RGBD. The results obtained are depicted in Tables \ref{table:comparisonSLAM} and \ref{table:comparisonSLAMrot} for translation and rotation, respectively.

The results in Tables \ref{table:comparisonSLAM} and \ref{table:comparisonSLAMrot} show that ORB-SLAM3 outperforms the other algorithms in most sequences. However, it fails to track those sequences with very low texture for Aqualoc and TUM-RGBD. As expected for direct SLAM algorithms, DSO fails under high parallax sequences caused by pure rotations and movements relative to close-up objects. Learning-based algorithms can infer under those harsh conditions, since they do not fail to execute under any of the sequences. However, the estimates in those sequences give large errors as a sign of wrong estimations.
It can be noted that the performance of learning-based algorithms is conditioned by the datasets under which they have been trained. In other words, they present better results for those sequences recorded under the same or similar conditions (i.e. KITTI and TUM-RGBD); however, the geometry-based algorithms outperform them in those sequences with very different motion patterns and environments. This is particularly noticeable in MIMIR-UW, which presents differently illuminated scenes. For EuRoC, TrianFlow presents lower relative errors to DF-VO. TrianFlow's model is trained under more \ac{DOF} than DF-VO, which may be the source of that difference. However, the hybrid architecture of DF-VO allows it to still provide a close performance to TrianFlow. In Aqualoc's case, the opposite is true, as the motions are closer to KITTI's, with lower errors for DF-VO.

While learning-based methods present a promising alternative in those visually-degraded conditions where geometry-based algorithms fail, it is necessary to assess the quality of the measurement, for instance, as an uncertainty value. On the other hand, geometry-based methods are still the prevailing techniques in those environments with a lack of data that do not allow training a reliable model.

The performance difference between geometry and learning-based approaches originates from the SLAM’s high dimensionality and limited data availability. The datasets like EuRoC and TUM-RGB are not large enough to train a network independently, whereas larger datasets like KITTI lack diversity in the data distribution.  The possibility of data augmentation is considered in Section \ref{sec:futureworks:trainingdata}. However, the high requirement for training data of learning-based SLAM originates from the
insufficient dimensionality of the pose regressors’ inductive bias, as discussed in Section \ref{sec:frontend:vo:supervised}. The approaches like DF-VO opt for hybrid approaches in pose regression to counteract this limitation. Another future direction would be implementing networks with embeddings representative of the geometric space, as discussed in Section \ref{sec:futureworks:generalizability}.

\begin{table*}[ht!]
\centering
\scriptsize
\caption{Results on the translation metrics obtained from deploying SLAM algorithms on the proposed datasets. The best results are marked in \textbf{bold}, and the second-best results are \underline{underlined}.}
\label{table:comparisonSLAM}
\resizebox{\textwidth}{!}{\begin{tabular}{|c|c|cc|cc|cc|cc|}
\hline
\multirow{2}{*}{Dataset}  & \multirow{2}{*}{Sequence}
& \multicolumn{2}{c|}{ORB-SLAM3 \cite{campos2021orb}} &  \multicolumn{2}{c|}{DSO \cite{engel2017dso}} & \multicolumn{2}{c|}{DF-VO \cite{zhan2019dfvo}}                     & \multicolumn{2}{c|}{TrianFlow \cite{zhao2020trianflow}}    \\ 
                                &            & APE[m] & RPE[m]  &  APE[m] & RPE[m] &  APE[m] & RPE[m] &  APE[m] & RPE[m]\\
\hline\hline

\multirow{4}{*}{KITTI \cite{dataset:kitti}} 
                                & 00       & \textbf{6.043} & \underline{0.1138}    &  115.67  & 4.23   & \underline{10.965}  & \textbf{0.0356} & 74.531  & 0.3415   \\
                                & 01       & \textbf{7.561} & \textbf{0.4616}    &  \underline{161.94}  & 18.53  & 201.253 & \underline{2.2723} & 256.990 & 2.6451   \\
                                & 02       & \underline{26.36} & \underline{0.2210}    &  89.72   & 2.98   & \textbf{17.285}  & \textbf{0.0417} & 129.550 & 0.3317   \\
                                & 03       & \underline{1.614} & \underline{0.0488}    &  136.99  & 1.62   & \textbf{0.589}  & \textbf{0.0220}  &  9.464  & 0.3157   \\
                                & 04       & \underline{1.386} & \underline{0.0893}    &  41.46   & 87.57  & \textbf{0.387}  & \textbf{0.0325}  &  3.162  & 0.2505  \\

\hline
\multirow{4}{*}{EuRoC \cite{burri2016euroc}} 
                                & MH\_01\_easy       & \textbf{0.0385} & \textbf{0.0285}    &  4.14  & 0.1449  & \underline{2.51}  & 0.0891   & 3.394  & \underline{0.031} \\
                                & MH\_04\_difficult  & \textbf{0.1043} & \underline{0.0745}    &  6.11  & 0.213   & \underline{3.498}  & 0.115   & 6.139  & \textbf{0.0575} \\
                                & V1\_02\_medium     & \textbf{0.0685} & \underline{0.0623}    &  \underline{1.750}  & 0.134   & 1.754  & 0.0664  & 1.760   & \textbf{0.0498}  \\
                                & V1\_03\_difficult  & \textbf{0.0761} & 0.0663    &  1.55  & 0.105   & \underline{1.372}  & \underline{0.0580}  & 1.531   & \textbf{0.0420} \\
\hline
\multirow{4}{*}{TUM-RGBD \cite{sturm2012tumrgbd}} 
                                & fr1/360       & \textbf{0.0069} & \textbf{0.0091}  & 0.179  & 0.046 & \underline{0.128}   & 0.0160 & 0.176  & \underline{0.0118} \\
                                & fr1/rpy       & \underline{0.0484} & 0.0310  &  -     & -     & \textbf{0.034 } & \underline{0.0050} & 0.0532 & \textbf{0.0042} \\
                                & fr3/nostructure\_notexture\_far   & -      & -       & -      & -     & \textbf{0.0634}  & \underline{0.0203} & \underline{0.561}  & \textbf{0.0183} \\
                                & fr3/nostructure\_texture\_near\_loop  & \textbf{0.0246} & \underline{0.0123}  & \underline{0.174}  & 0.020 & 0.628   & 0.0190 & 1.729  & \textbf{0.0110} \\

\hline
\multirow{4}{*}{Aqualoc \cite{dataset:aqualocdb}} 
                                & Archaeo 1   & \textbf{0.0124} & \textbf{0.0138}  & 2.60 & 4.16 &  \underline{2.12}   & \underline{0.037} & 2.57 &  2.39  \\
                                & Archaeo 2   & \textbf{0.0280 }& \textbf{0.0347}  & 5.31 & 7.01 &  4.12   & \underline{0.180} & \underline{2.65} &  5.14  \\
                                & Archaeo 3   &\textbf{ 0.006}  & \textbf{0.0058} & 3.36 & 3.61 &  \underline{1.088}  & \underline{0.057} & 1.13 &  1.68  \\
                                & Archaeo 4   & -      &  -      & 3.13 & 4.36 &  \textbf{0.248}  & \textbf{0.111} & \underline{0.267} & \underline{0.327} \\

\hline
\multirow{4}{*}{MIMIR-UW \cite{dataset:mimir}} 
                                & SeaFloor/0       & \textbf{3.67} & 0.13 &  - & - &  \underline{13.24}  & 0.143 & 16.04 &  \textbf{0.123}  \\
                                & SeaFloor/1       & \underline{8.78} & 0.19  & \textbf{2.73} & \textbf{0.0053} &  13.99  & 0.127 & 18.65 &  \underline{0.126}  \\
                                & SeaFloor\_Algae/1     & \textbf{1.15} & 0.134   & \underline{7.00} & \textbf{0.048}   & 14.083  & \underline{0.120} & 16.61 & 0.125  \\
                                & OceanFloor/0\_dark & \textbf{5.78} & \underline{0.523}   &  - & - &  \underline{17.649}  & \textbf{0.144} & 17.90 &  0.544  \\

\hline

\end{tabular}}
\end{table*}

\begin{table*}[ht!]
\centering
\scriptsize
\caption{Results on the rotation metrics obtained from deploying SLAM algorithms on the proposed datasets. The best results are marked in \textbf{bold}, and the second-best results are \underline{underlined}.}
\label{table:comparisonSLAMrot}
\resizebox{\textwidth}{!}{\begin{tabular}{|c|c|cc|cc|cc|cc|}
\hline
\multirow{2}{*}{Dataset}  & \multirow{2}{*}{Sequence}
& \multicolumn{2}{c|}{ORB-SLAM3 \cite{campos2021orb}} &  \multicolumn{2}{c|}{DSO \cite{engel2017dso}} & \multicolumn{2}{c|}{DF-VO \cite{zhan2019dfvo}}                     & \multicolumn{2}{c|}{TrianFlow \cite{zhao2020trianflow}}    \\ 
                                &            & APE[rad] & RPE[rad]  &  APE[rad] & RPE[rad] &  APE[rad] & RPE[rad] &  APE[rad] & RPE[rad]\\
\hline\hline

\multirow{4}{*}{KITTI \cite{dataset:kitti}} 
                                & 00       & \textbf{0.0207} & \underline{0.0043} & 1.103 & 0.2588 & \underline{0.0532} & \textbf{0.0019} & 0.8790 & 0.0278 \\
                                & 01       & \textbf{0.0165} & \textbf{0.0019} & 0.7180 & 0.1941   & 2.8907  & 0.0995 & \underline{0.3903} & \underline{0.0023}  \\
                                & 02       & \textbf{0.0275} & \underline{0.0020}  &  0.3814  & 0.1280   & \underline{0.0685}  & \textbf{0.0011} & 0.6706  & 0.0136  \\
                                & 03       & \textbf{0.0082} & \underline{0.0009}    & 1.7836 & 0.1045   & \underline{0.0298}  & \textbf{0.0008} & 0.2579  & 0.0234  \\
                                & 04       & \underline{0.0878} & \underline{0.0008}    &  1.909  & 0.0052   & \textbf{0.0644}  & \textbf{0.0005} & 0.1403 & 0.0020  \\
\hline
\multirow{4}{*}{EuRoC \cite{burri2016euroc}} 
                                & MH\_01\_easy       & \underline{1.557} & \textbf{0.0143} &  1.940  & 0.0642 & \textbf{1.549}  & \underline{0.0182} & 2.007  & 0.0205  \\
                                & MH\_04\_difficult  & \textbf{1.550} & \textbf{0.0165}  &  1.745  & 0.057  & \underline{1.585}  & \underline{0.0205} & 1.992  & 0.0221  \\
                                & V1\_02\_medium     & \textbf{1.549} & 0.0519  &  2.429  & 0.1189  & \underline{2.276}  & \underline{0.0484} & 2.4497  & \textbf{0.0407} \\
                                & V1\_03\_difficult  & \textbf{1.542} & 0.0757  &  2.295  & 0.1255   & 2.109  & \underline{0.0554} & \underline{1.763}  & \textbf{0.0472}  \\
\hline
\multirow{4}{*}{TUM-RGBD \cite{sturm2012tumrgbd}} 
                                & fr1/360                               & \textbf{0.2684} & \textbf{0.0062}  &  2.2604  & 0.1600 & \underline{1.875}  & 0.3036 & 2.093  & \underline{0.0794}  \\
                                & fr1/rpy                               & \textbf{1.821} & \textbf{0.0260} &  - & - & 2.739 & 0.2160 & \underline{2.600} & \underline{0.0689}  \\
                                & fr3/nostructure\_notexture\_far       & -      & -       & -      & -     & \textbf{2.091}  & \underline{0.0321} & \underline{2.838}  & \textbf{0.0268} \\
                                & fr3/nostructure\_texture\_near\_loop  & \textbf{0.0248} & \textbf{0.0133} &  \underline{0.111}  & 0.0335  & 1.843  & 0.0472 & 2.285  & \underline{0.0219} \\

\hline
\multirow{4}{*}{Aqualoc \cite{dataset:aqualocdb}} 
                                & Archaeo 1   & \textbf{0.0197} & \textbf{0.0145}  &  2.659  & 1.919   & \underline{2.382}  & \underline{0.0267} & 2.459  & 1.382  \\
                                & Archaeo 2   & \textbf{0.0413} & \textbf{0.0190} & \underline{0.6874} & 0.8024  & 2.491 & \underline{0.2630} & 2.113  & 1.852  \\
                                & Archaeo 3   & \textbf{0.0492} & \textbf{0.0292} &  \underline{0.6797}  & 0.9806  & 2.117  & \underline{0.264} & 2.342  & 2.12  \\
                                & Archaeo 4   & -      &  -      & \textbf{0.5936} &  \underline{0.7571} & \underline{1.780} & \textbf{0.2870} & 2.393 & 1.294\\
\hline
\multirow{4}{*}{MIMIR-UW \cite{dataset:mimir}} 
                                & SeaFloor/0       & 2.344 & \textbf{0.0191} &  -  & -   &  \textbf{1.959}  & \underline{0.0203} & \underline{2.069} & 0.0256  \\
                                & SeaFloor/1       & 2.472 &  0.0241 & 2.783 & 0.0700 &  \textbf{2.028}  & \textbf{0.0170} & \underline{2.445} & \underline{0.0223} \\
                                & SeaFloor\_Algae/1     & \underline{2.339} & 0.0249 & 2.827 & 1.129 &  \textbf{2.010}  & \textbf{0.0173} & 2.559 & \underline{0.0217} \\
                                & OceanFloor/0\_dark & \underline{2.073} & \textbf{0.0078}   &  - & - &  2.474  & 0.1696 & \textbf{2.029} &  \underline{0.0660}  \\

\hline

\end{tabular}}
\end{table*}
\section{Open problems and future directions}
Monocular visual SLAM comprises a complex and heterogeneous algorithm. Decades of development of geometry-based solutions have led to a set of accurate and efficient algorithms that nowadays make geometry-based SLAM the de-facto standard in robotics. The following section discusses a series of open problems and suggests future directions to bring deep learning-based implementations closer to becoming the new standard for SLAM.

\subsection{Training data}
\label{sec:futureworks:trainingdata}
When it comes to training models, the amount of data available can significantly impact accuracy and reproducibility. Data augmentation techniques are often used to feed more data to the networks. However, this can make the training process challenging to replicate. The reproducibility can be improved using standard data splits, as done with KITTI \cite{kittieigensplit}. Additionally, it is important to ensure unbiased data distributions to train more accurate models. This can be achieved by quantifying any biases and designing a data split that minimizes them.

\subsection{Towards generalizable models}
\label{sec:futureworks:generalizability}
Alleviating the biases in the data distribution using data augmentation techniques has proven to improve generalizability in models in the literature. However, the next step towards achieving even greater results is the development of architectures that can learn in higher dimensions.
Pose regression modules often rely on classification networks with a shift-equivariant inductive bias. To make models more generalizable, one potential future work is the implementation of networks with higher-dimensional inductive biases. Section \ref{sec:deeplearning} presents a set of networks that meet this requirement.
GNNs, for example, have already been successfully integrated into end-to-end loop detection (see Section \ref{sec:deeplearning:loop}), showing potential for being integrated into other modules of the SLAM's architecture. 
Additionally, another important aspect of generalization is accounting for the camera's intrinsic parameters. By doing so, models can better account for a wider range of setups and include more data from different sources during training, thus improving the overall data's distribution.

\subsection{Closing the loop: computational expense}
It is well known that visual SLAM challenges the computer's computational power where it is leveraged. In geometry-based SLAM, the optimization from the front-end and the back-end, and the loop search from the loop detection algorithm, are three demanding algorithms that run simultaneously.
Deep learning implementations measure the progress of these algorithms in terms of accuracy, with less focus on computational expense. Thus, executing them within a complete SLAM pipeline is often unsuitable, one of the main open problems in deep learning-based SLAM. Geometry-based pipelines have proved that trading off the accuracy on the front-end with lower computational expense, in the end, leverages better results by allowing the optimization of the estimates in the back-end. Similarly, this realization could be imported into deep learning pipelines, integrating less accurate and computationally expensive models with a SLAM back-end. One way to face this could be by leveraging a pose graph optimization process that merely optimizes the poses and closes loops, as introduced in Section \ref{sec:backend:loopclosingandgraphopt}. Another way is to leverage a bundle adjustment algorithm. This brings up other challenges: integrating the model's latent space and output estimates into a graph and designing a cost function for optimizing them.

\subsection{From end-to-end front-end to end-to-end SLAM}
An alternative direction to the one presented before would be the development of a complete architecture for SLAM in an end-to-end fashion. Section \ref{sec:deeplearning:endSLAM} presents some preliminary works in this area. One deep learning architecture that exhibits promise for this direction is \acp{GNN}. GNNs are capable of leveraging inductive bias in higher dimensions,  and they enable the integration of heterogeneous data such as images and poses. Furthermore, exploiting the graph's sparsity brings the potential for computationally efficient models.

\subsection{Not just visual: model flexibility}
This survey primarily addresses monocular visual SLAM methods, but it is important to mention that geometry-based frameworks have the potential to incorporate other sensor measurements. By combining external measurements with the graph, these frameworks can produce better estimates from both data sources. However, integrating external measurements into deep learning pipelines is more challenging. One possible future direction is exploring the development of architectures that can handle a broader range of data sources.

\subsection{Reliability}
Challenging imaging conditions lead geometry-based algorithms to lose track and deep learning-based algorithms to retrieve spurious estimates. Reliable algorithms must be able to assess and handle the occurrence of such events.
In geometry-based SLAM, a loss of tracking can be readily identified by the inability to make an estimate owing to inconsistent or inadequate data. Conversely, deep learning models always retrieve an estimate.
Thus, an emerging field of study for deep learning networks is incorporating an uncertainty estimate into the output.

\section{Conclusions and future work}
\label{sec:futurework}
This paper aims to provide an overview of monocular visual SLAM, formulating the different classes into which it is categorised according to its taxonomy, and the algorithms that comprise it. First, geometry-based SLAM algorithms are presented to constitute the front and back-end. Later, deep learning pipelines are elaborated, which have led to the development of end-to-end methods. We believe the deep learning-based end-to-end SLAM algorithms set up a new paradigm for traditional SLAM.

Geometry-based monocular SLAM presents a well-established solution in terms of significant efficiency. The experiments showcase its known limitations: track failure and drift under visually or geometrically degraded scenes. Deep learning models tackle those limitations with higher-level representations of the environment, but the training environments limit their performance.
This limitation is mainly addressed by feeding more and more varied data to the network. However, considering the wide variety of deployment conditions, it is - for most of the applications - likely insufficient. It is then desired to achieve generalizability from the network's design side and not just from the data fed to it. Recently, studies on the underlying geometry under deep network architectures have arisen. The main findings through the comparison tests in this review for deep learning-based SLAM show that applying  geometric deep learning
pipelines remains still an open problem. This, together with the advances in fully-differentiable frameworks, and the study of continuous presentations for SLAM, presents promising lines of development that can meet a compromise between geometry and learning-based pipelines.

\ifCLASSOPTIONcaptionsoff
  \newpage
\fi

\bibliographystyle{IEEEtran}
\bibliography{References}

\begin{IEEEbiography}
[{\includegraphics[width=1in,height=1.25in,clip=true]{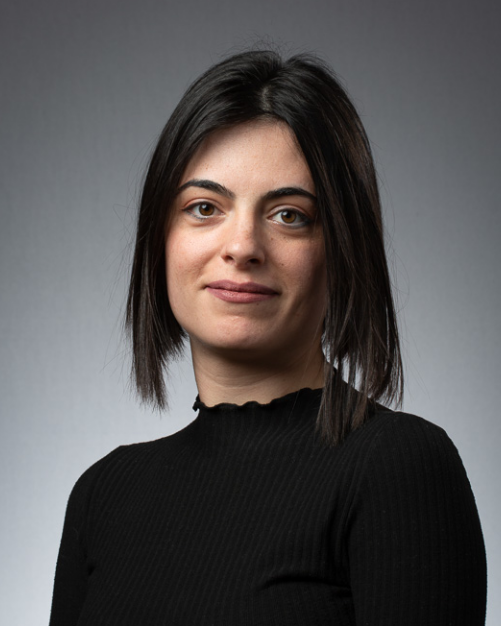}}]{Olaya Álvarez Tuñón}
(Student Member) is a Ph.D. student at Aarhus university. She holds a degree in Electronics Engineering (2015) and an MSC in Robotics and Automation (2019) from the University Carlos III of Madrid. She worked as a research assistant in the RFCS project Long-term STability Assessment and Monitoring of flooded Shafts (STAMS). Her current research is framed within the European project Reliable AI for Marine Robotics (REMARO). Within REMARO, she works in the development of algorithms for vision-based navigation for underwater safety-critical applications.
\end{IEEEbiography}

\begin{IEEEbiography}    [{\includegraphics[width=1in,height=1.25in,clip,keepaspectratio]{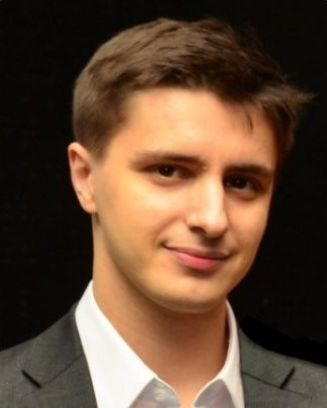}}]{Yury Brodskiy} received his Ph.D. from the University of Twente, The Netherlands. His research interests include  control for underwater AUV, interaction control for mobile manipulators, as well as classical and deep-learned visual underwater SLAM. He is a senior software developer and team lead for the AI vision group at Eiva A/S.  
\end{IEEEbiography}

\begin{IEEEbiography}    [{\includegraphics[width=1in,height=1.25in,clip,keepaspectratio]{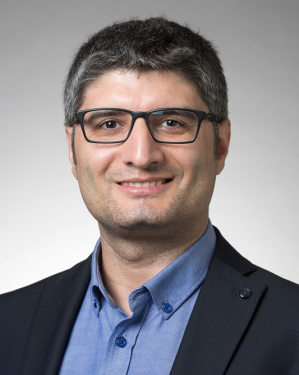}}]{Erdal Kayacan}
(Senior Member) received a PhD degree in electrical and electronic engineering from Bogazici University, Istanbul, Turkey, in 2011. After finishing his postdoctoral research with the University of Leuven (KU Leuven), Leuven, Belgium, in 2014, he worked with the School of Mechanical and Aerospace Engineering, Nanyang Technological University (NTU), Singapore, as an assistant professor for four years. He was an associate professor at Aarhus University at the Department of Engineering from 2018 to 2023. He is pursuing his career as a full professor at the Department of Electrical and Information Technology at Paderborn University, Germany.
His research areas are computational intelligence methods, sliding mode control, model predictive control, mechatronics and unmanned aerial vehicles.

Dr Kayacan is a member of the Computational Intelligence Society and the Robotics and Automation Society in the Institute of Electrical and Electronics Engineers (IEEE).  He is the IEEE/ASME Transactions Mechatronics technical editor and the associate editor in IEEE Robotics and Automation Letters.
\end{IEEEbiography}


\end{document}